\documentclass[10pt,twocolumn,letterpaper]{article}

\usepackage{nicefrac}
\usepackage{subfigure}
\usepackage[table]{xcolor}
\usepackage{multirow}
\usepackage{amssymb}
\usepackage{amsfonts}
\usepackage{booktabs}
\usepackage{adjustbox,lipsum}
\usepackage{cvpr}
\usepackage{times}
\usepackage{epsfig}
\usepackage{graphicx}
\usepackage{amsmath}
\usepackage{amssymb}
\usepackage{verbatim}
\usepackage{makecell}
\usepackage{balance}
\usepackage[pagebackref=true,breaklinks=true,letterpaper=true,colorlinks,bookmarks=false]{hyperref}

 \cvprfinalcopy % *** Uncomment this line for the final submission

\def\cvprPaperID{4334} % *** Enter the CVPR Paper ID here

\ifcvprfinal\fi
\begin{document}

%%%%%%%%% TITLE
%\title{Spherical Convolutional Neural Network \\ for 3D Point Clouds}

%\title{Spherical Convolutional Neural Network  for 3D Point Clouds}
\title{Octree guided CNN with Spherical Kernels for 3D Point Clouds}
\author{Huan Lei\hspace{6mm}Naveed Akhtar\hspace{6mm}Ajmal Mian\\
School of Computer Science and Software Engineering\\
The University of Western Australia\\
{\tt\small huan.lei@research.uwa.edu.au, \{naveed.akhtar,ajmal.mian\}@uwa.edu.au}
}

\maketitle
%\thispagestyle{empty}

%%%%%%%%% ABSTRACT
\begin{abstract}
\vspace{-3mm}
   We propose an octree guided neural network architecture and spherical convolutional kernel for machine learning from arbitrary 3D point clouds. The network architecture capitalizes on the sparse nature of irregular point clouds, and hierarchically coarsens the data  representation with space partitioning. At the same time, the proposed  spherical kernels systematically quantize point neighborhoods to identify local geometric structures in the data, while maintaining the properties of translation-invariance and asymmetry. We specify spherical kernels with the help of network neurons that in turn are associated with spatial locations. We exploit this association to avert dynamic kernel generation during network training that enables efficient learning with high resolution point clouds. The effectiveness of the proposed technique is established on the benchmark tasks of 3D object classification and segmentation, achieving new state-of-the-art on ShapeNet and RueMonge2014 datasets. 
   
 %  Our metric-based s. The network architecture itself is guided by octree data structuring that  takes full advantage of the sparse nature of irregular point clouds, and hierarchically coarsens the representation. We specify spherical
 \end{abstract}

\vspace{-5mm}
\section{Introduction}
\vspace{-2mm}
Convolutional Neural Networks (CNNs) \cite{lecun1998gradient} are known
to learn highly effective features from data.
However, standard CNNs are only amenable to data defined over regular grids, e.g.~pixel arrays. This  limits their ability in processing 3D point clouds that are inherently irregular. 
Point cloud processing has recently gained significant research interest and large repositories for this data modality have started to emerge \cite{armeni20163d, chang2015shapenet, hackel2017semantic3d, wu20153d, yi2016scalable}.
Recent literature has also seen many attempts to exploit the representation prowess of standard convolutional networks for point clouds by adaption \cite{maturana2015voxnet, wu20153d}.
However, these attempts have often led to excessively large memory footprints that restrict the allowed input data resolution \cite{riegler2017octnet, simonovsky2017dynamic}.
A more attractive choice is to combine the power of convolution operation  with graph representations of irregular data.
The resulting Graph Convolutional Networks (GCNs) offer convolutions either in  spectral domain  \cite{bruna2013spectral, defferrard2016convolutional,kipf2017semi} or  spatial domain \cite{simonovsky2017dynamic}.

In GCNs, the spectral domain methods require the Graph Laplacian to be aligned, which is not straight forward to achieve for point clouds. On the other hand, the only prominent approach in  spatial domain is the Edge Conditioned filters in CNNs for graphs (ECC)~\cite{simonovsky2017dynamic} that, in contrast to the standard CNNs, must generate convolution kernels dynamically entailing a significant computational overhead. Additionally, ECC relies on range searches for graph construction and coarsening, which can become prohibitively expensive for large point clouds. 
One major challenge in applying convolutional networks to irregular 3D data is in specifying geometrically meaningful convolution kernels in the 3D metric space.
Naturally, such kernels are also required to exhibit translation-invariance to identify similar local structures in the data. Moreover, they should be applied to point pairs asymmetrically for a compact representation.   
Owing to such intricate requirements, few existing techniques   altogether avoid the use of convolution kernels in computational graphs to process  unstructured data \cite{klokov2017escape, qi2017pointnet, qi2017pointnetplusplus}. Although still attractive, these methods do not contribute towards harnessing the potential of convolutional neural networks for point clouds. 

\begin{figure*}[t]
  \centering
  \includegraphics[height=43mm]{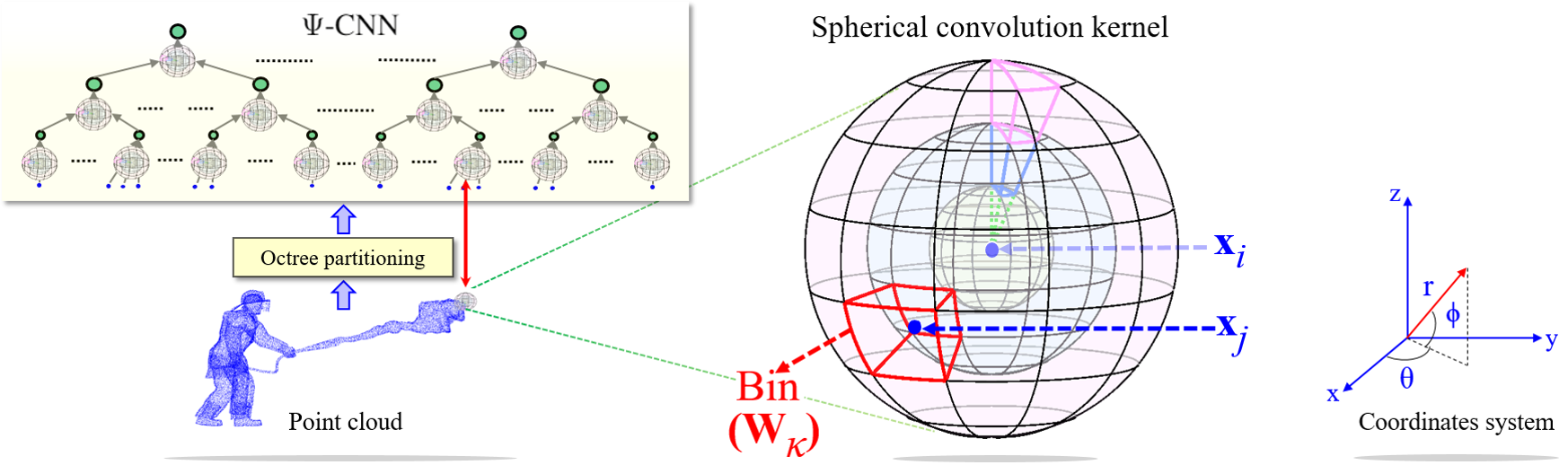}
  \caption{The proposed octree guided CNN, i.e.~$\Psi$-CNN directly processes raw point clouds using octree partitioning information. The representation is  hierarchically coarsened at each network layer (three layers depicted) by applying spherical convolutional kernels. A spherical kernel systematically splits the space around a  point ${\bf x}_i$ into multiple volumetric bins. For the $j^{\text{th}}$ neighboring point ${\bf x}_j$, a kernel first determines its relevant bin and uses the weight matrix ${\bf W}_{\kappa}$ defined for that bin to compute the  activation value. The proposed spherical kernel preserves translation-invariance and asymmetry properties of standard 2D convolutional kernel in 3D point cloud domain.  }
  %\label{sphConv}
  \label{fig:sphConv1}
  \vspace{-4mm}
\end{figure*}

In this work, we introduce the notion of spherical convolutional kernel that systematically partitions a spherical 3D region into multiple volumetric bins, see Fig.~\ref{fig:sphConv1}. Each bin of the kernel specifies a matrix of learnable parameters that weights the points falling within that bin for convolution.
We apply these kernels between the layers of a Neural Network ($\Psi$-CNN) that we propose to construct by exploiting octree partitioning~\cite{meagher1982geometric} of the 3D space. The sparsity guided octree structuring determines the locations to perform the convolutions in each layer of the network. The network architecture itself is guided by the hierarchy of the octree, having the same number of hidden layers as the tree depth. By exploiting space partitioning, the network avoids K-NN/range search and efficiently consumes high resolution point clouds. 
It also avoids dynamic generation of the proposed kernels by associating them to its neurons. At the same time, 
%Training of our network does not require dynamic generation of the proposed kernels, whereas 
the kernels are able to share weights between similar local structures in the data. We theoretically establish that the spherical kernels are applied asymmetrically to points in our network just as the kernels in standard CNNs are applied asymmetrically to image pixels. This ensures compact representation learning by the proposed network in the point cloud domain.  
We demonstrate the effectiveness of our method for 
3D object classification, part segmentation and large-scale semantic segmentation. 
% With a novel concept of convolutional networks, We achieve state-of-the-art results on standard segmentation benchmarks, and
% competitive results for the classification benchmarks.
The major contributions of this work are summarized  below:
\begin{itemize}
\item A novel concept of translation-invariant and asymmetric convolutional kernel is proposed and analyzed for point-wise feature learning from irregular point clouds. 
%A spherical convolutional kernel, which is translation-invariant and asymmetric, is introduced to learn point-wise features for the point cloud in an irregular way.  It exploits the intrinsic sparsity of point clouds.
\item The resulting convolutional kernel is exploited with an octree guided neural network that, in contrast to the previous voxelization applications of octree to point clouds~\cite{riegler2016octnet}, hierarchically coarsens the data and constructs point neighborhoods using space partitioning to avoid time-consuming K-NN/range search.
\item Efficacy of the proposed technique is established by experiments with ModelNets~\cite{wu20153d} for 3D object classification, ShapeNet~\cite{yi2016scalable} for part segmentation, and  RueMonge2014~\cite{riemenschneider2014learning} for semantic segmentation, achieving new state-of-the art on the last two datasets.   
\end{itemize}

\section{Related Work}
PointNet~\cite{qi2017pointnet} is one of the first instances of exploiting neural networks to represent point clouds. It directly uses $x,y,z$-coordinates of points as input features. The network learns point-wise features with shared MLPs, and extracts a global feature with max pooling. A major limitation of PointNet is that it explores no geometric context in point-wise feature learning. This was later addressed by PointNet++~\cite{qi2017pointnetplusplus} with hierarchical application of max-pooling to the local regions.  The enhancement builds local regions  using K-NN search as well as range search. Nevertheless, both PointNets \cite{qi2017pointnet,qi2017pointnetplusplus} aggregate the  context information with max pooling, and no convolution modules are explored in the networks. 
In regards to processing point clouds with deep learning  using tree structures, Kd-network~\cite{klokov2017escape} is among the pioneering prominent contributions. Kd-network also uses point coordinates as its input, and computes feature of a parent node by concatenating the features of its children in a \emph{balanced} tree. However, its performance depends heavily on the randomization of the tree construction. This is in sharp contrast to our approach that uses deterministic geometric relationships between the points. 
Another technique, SO-Net \cite{li2018so} reorganizes the irregular point cloud into an $m\times m$ 2D rectangular map, and uses the PointNet architecture to learn
node-wise features for the map. Similarly, KCNet \cite{shen2018mining} also builds on PointNet and 
introduces a point-set template to learn geometric correlations of local points in the point cloud.  PointCNN~\cite{li2018pointcnn} extracts permutation-invariant features by reordering the local points canonically with a learnable $\chi$-transformation. 
All of these methods relate to our work in terms of directly accepting the spatial coordinates of points as  input. However, they do not contribute towards the use of  convolutional networks for processing 3D point clouds. Approaches advancing that research direction can be divided into two broad categories, discussed below.  

%The major similarity between our approach and the above mentioned approaches comes in terms of directly accepting the spatial coordinates of points as the input features. From the perspective of convolutional networks for 3D data, the following two categories of networks also  relate to our approach.

\vspace{-2mm}
\subsection{Graph Convolutional Networks}
\vspace{-2mm}
Graph convolutional networks can be grouped into spectral networks \cite{bruna2013spectral,defferrard2016convolutional,kipf2017semi} and spatial networks \cite{simonovsky2017dynamic}. The spectral networks perform convolution in the spectral domain relying on the graph Laplacian and adjacency matrices, while the spatial networks perform convolution in the spatial domain. 
A major limitation of spectral networks is that they demand the graph structure to be fixed, which makes their application to the data with varying graph structures (e.g.~point clouds) challenging. Yi \emph{et al.} \cite{yi2017syncspeccnn} attempted to address this issue with Spectral Transformer Network (SpecTN), similar to STN \cite{jaderberg2015spatial} in the spatial domain. However, the signal transformation from spatial to spectral domains and vice-versa results in  computational complexity $\mathcal{O}(n^2)$. 
ECC~\cite{simonovsky2017dynamic} is among the pioneering works  for point cloud analysis with graph convolution in the spatial domain. Inspired by the dynamic filter networks \cite{de2016dynamic}, it adapts MLPs to generate convolution filters between the connected vertices dynamically. 
 The dynamic generation of filters comes with computational overhead. Additionally, the neighborhood construction and graph coarsening in ECC must rely on range searches, which is not efficient. We achieve  
coarsening and neighborhood construction directly from the  octree partitioning, thereby avoiding expensive range searching. Moreover, our spherical convolutional kernel effectively explores the geometric context of each point without requiring dynamic filter generation.

\vspace{-1mm}
\subsection{3D Convolutional Neural Networks} 
\vspace{-2mm}
3D-CNNs are  applied to volumetric representations of 3D data. 
 In the earlier attempts in this direction, only low input resolution  could be processed, e.g.~30$\times$30$\times$30 \cite{wu20153d}, 32$\times$32$\times$32 \cite{maturana2015voxnet}.
This issue transcended to subsequent works as well~\cite{huang2016point,sedaghat2016orientation,zeng20163dmatch,zhang2017deepcontext}. 
The limitation of low input resolution was a natural consequence of the cubic growth of memory and computational requirements associated with the volumetric input data. 
Later methods~\cite{EngelckeICRA2017,li2016fpnn} mainly aim at addressing these issues. 
Most recently, 
Riegler \emph{et al.}~\cite{riegler2017octnet} proposed OctNet, that represents point clouds with a hybrid of shallow grid octrees (depth=3). Compared to its dense peers, OctNet reduces the computational and memory costs to a large degree, and is applicable to high-resolution inputs up to 256$\times$256$\times$256. 
Whereas OctNet also exploits octrees, there are major differences between OctNet and our method. 
Firstly, OctNet must process point clouds as regular 3D volumes due to its 3D-CNN kernels. No such constraint is applicable to our technique due to the proposed  spherical kernels. Secondly, we are able to learn point cloud representation with a single deep octree instead of using hybrid of shallow trees. %{\color{red}This results from... }

%{\color{blue} Although our architecture is based on octree as well, the proposed spherical convolution differs our work significantly from OctNet. 
%Firstly, the 3D-CNN kernel determines that 
%OctNet processes the point cloud as regular 3D volumes, while we process the point cloud following its irregular nature. Secondly, our network exploits 
% $x,y,z$ coordinates of points as features while OctNet does not. In addition, our network is adapted from a single deep octree rather than a hybrid of shallow octrees, as in OctNet.}
%-------------------------------------------------------------------------
\section{Spherical Convolutional Kernel} \label{sec:SC}
\vspace{-1mm}
Our network derives its main strength from spherical convolutional kernels. Thus, it is imperative to first understand the proposed kernel before delving into the network details. This section explains our convolutional kernel for 3D point cloud processing.     

For images, hand-crafted features have traditionally been computed over more primitive constituents, i.e.~patches. In effect, the same principle transcended to automatic feature extraction with the standard CNNs that compute feature maps using the activations of well-defined rectangular regions. Whereas rectangular regions are a common choice to process data of 2D nature, spherical regions are more suited to process unstructured 3D data such as point clouds. Spherical regions are inherently amenable to computing geometrically meaningful features for such data \cite{frome2004recognizing,tombari2010unique,tombari2010uniqueACM}.
%Such regions also provide an intuitive and systematic partitioning of the 3D space.
Inspired by this natural kinship, we introduce the concept of \textit{spherical convolutional kernel}\footnote{Note that the term \emph{spherical} in Spherical CNN~\cite{cohen2018spherical} is used for spherical surfaces (i.e.~$360^\circ$ images) not the ambient 3D space. Our concept of spherical kernel widely differs from~\cite{cohen2018spherical}, and it is used in different context.} that uses a 3D sphere as the basic geometric shape to perform the convolution. 

Given an arbitrary point cloud $\mathcal{P}=\{\mathbf{x}_i\in \mathbb{R}^3\}_{i=1}^m$, where $m$ is the number of points; we define the convolution kernel with the help of a sphere of radius $\rho\in \mathbb{R}^+$. For a target point $\mathbf{x}_i$, we consider its neighborhood $\mathcal{N}(\mathbf{x}_i)$ to comprise the points within the sphere centered at $\mathbf{x}_i$,  i.e. $\mathcal{N}(\mathbf{x}_i)=\{\mathbf{x}:d(\mathbf{x},\mathbf{x}_i)\leq \rho\}$, where $d(.,.)$ is a distance metric - $\ell_2$ distance in this work.
We divide the sphere into $n \times p \times q$ \emph{`bins'} (see Fig.~\ref{fig:sphConv1}) by partitioning the occupied space uniformly along the azimuth ($\theta$) and elevation ($\phi$) dimensions. We allow the partitions along the radial dimension  to be non-uniform because cubic volume growth for large radius values can become undesirable.
Our quantization of the spherical region is mainly inspired by  3DSC \cite{frome2004recognizing}.
%The sphere is divided into many bins along its azimuth, elevation and radial dimensions, similar to the division scheme used in 3DSC.
We also define an additional bin corresponding to the origin of the sphere to allow the case of self-convolution of points.
%{\color{blue}Besides,
%  to allow convolution on each point itself, we also define
 % an additional bin for $d(\bold{x}_i,\bold{x}_i)=0$.
%In this case, there will be no splitting along the azimuth and elevation directions.}
For each bin, we define a weight matrix $\mathbf{W}_{\kappa \in \{0, 1,\dots,n \times p \times q \}}\in\mathbb{R}^{s\times t}$ of learnable parameters, where $s$-$t$ are the number of output-input channels and $\mathbf{W}_0$ relates to self-convolution.  %{\color{red}(Need some intuition here about the significance/meaning of weights, e.g. how they relate to standard convolution weights, their dimensions ?)}.
Together, the $n \times p \times q + 1$ weight matrices specify a single spherical convolutional kernel.

%$(n,p,q\in \mathbb{Z}^+), (s,t\in \mathbb{Z}^+)$

%--------------------------------------------------------------------
To compute the activation value for a target point $\mathbf{x}_i$, we must identify the relevant weight matrices of the kernel for each of its neighboring points $\mathbf{x}_j\in \mathcal{N}(\mathbf{x}_i)$.
It is straightforward to associate $\mathbf{x}_i$ with $\mathbf{W}_0$ for self-convolution. For the non-trivial cases, we first represent the neighboring points in terms of their spherical coordinates that are referenced using $\mathbf{x}_i$ as the origin. That is, for each $\mathbf{x}_j$ we compute $\mathcal T(\boldsymbol{\Delta}_{ji})\rightarrow \boldsymbol{\psi}_{ji}$, where $\mathcal T(.)$ defines the transformation from Cartesian to Spherical coordinates and  $\boldsymbol{\Delta}_{ji}=\mathbf{x}_j-\mathbf{x}_i$.
Assuming that the bins of the quantized sphere are indexed by $k_\theta$, $k_\phi$ and $k_r$ along the azimuth, elevation and radial dimensions respectively, the weight matrices associated with the spherical kernel bins can be indexed as $\kappa = k_\theta + (k_\phi-1)\times n + (k_r-1)\times n\times p$, where $k_\theta \in \{1,\dots,n\},~k_\phi \in \{1,\dots,p\},~k_r \in \{1,\dots,q\}$. Using this indexing, we assign  each $\boldsymbol{\psi}_{ji}$; and hence $\mathbf{x}_j$ to its relevant weight matrix.
In the $l^{\text{th}}$ network layer, the activation for the $i^{\text{th}}$ point can then be computed as:
\begin{align}\label{spatio-conv}
  &\bold{z}^l_i  = \frac{1}{|\mathcal{N}(\bold{x}_i)|}\sum\limits_{j = 1}^{|\mathcal{N}(\bold{x}_i)|}\bold{W}^l_{\kappa}\bold{a}^{l-1}_j+\bold{b}^l,\\
  &\bold{a}^l_i = f(\bold{z}^l_i),
  \end{align}
where $\bold{a}^{l-1}_j$ is the activation value of a neighboring point from layer $l-1$, $\bold{W}^l_{\kappa}$ is the weight matrix, $\bold{b}^l$ is the bias vector, and $f(\cdot)$ is the non-linear activation function - ReLU~\cite{nair2010rectified} in our experiments.

To elaborate on the characteristics of the proposed spherical convolutional kernel, let us denote the \emph{edges} of the kernel bins along $\theta$, $\phi$ and $r$ dimensions respectively as:
%Let the bin edges for $\theta$, $\phi$ and $r$ be
\begin{align}
\notag
&\boldsymbol{\Theta}=[\Theta_1,\dots,\Theta_{n+1}], ~\Theta_k<\Theta_{k+1}, {\Theta}_k\in[-\pi,\pi], \\
\notag
&\boldsymbol{\Phi}=[\Phi_1,\dots, \Phi_{p+1}]\big],~\Phi_k<\Phi_{k+1}, {\Phi}_k\in\big[-\frac{\pi}{2}, \frac{\pi}{2}], \\
\notag
&\bold{R}=[R_1,\dots,R_{q+1}], ~~R_k<R_{k+1}, R_k\in(0,\rho].
\end{align}
Due to the constraint of uniform splitting along the azimuth and elevation, we can write  ${\Theta}_{k+1}-{\Theta}_k=\frac{2\pi}{n}$ and ${\Phi}_{k+1}-{\Phi}_k=\frac{\pi}{p}$.% and $R_{k+1}-R_k=\frac{\rho}{q}$.

\noindent {\bf Lemma 2.1:} \emph{If~$\Theta_k\cdot\Theta_{k+1}\geq 0$,  $\Phi_k\cdot\Phi_{k+1}\geq 0$ and $n>2$, then for any two points $\bold{x}_a\neq\bold{x}_b$ within the spherical convolutional kernel, the weight matrices $\mathbf W_{\kappa}, \forall \kappa>0$, are applied asymmetrically.}\\
\noindent {\it Proof:}
% Let $\bold{x}_a$ and $\bold{x}_b$ be two different points within the spherical convolution kernel. Then we have $\bold{x}_a\in\mathcal{N}(\bold{x}_b)$ and $\bold{x}_b\in\mathcal{N}(\bold{x}_a)$.
Let $\boldsymbol{\Delta}_{ab}= {\bf x}_a - {\bf x}_b =[\delta_x,\delta_y,\delta_z]^{\intercal}$, then $\boldsymbol{\Delta}_{ba}=[-\delta_x,-\delta_y,-\delta_z]^{\intercal}$.  Under the Cartesian to Spherical coordinate transformation,  we have $\mathcal T(\boldsymbol{\Delta}_{ab}) =\boldsymbol{\psi}_{ab}=[\theta_{ab},\phi_{ab},r]^{\intercal}$,  and $\mathcal T(\boldsymbol{\Delta}_{ba}) = \boldsymbol{\psi}_{ba}=[\theta_{ba},\phi_{ba},r]^{\intercal}$. We assert that $\boldsymbol{\psi}_{ab}$ and $\boldsymbol{\psi}_{ba}$ fall in the same bin indexed by $\kappa \leftarrow (k_\theta,k_\phi,k_r)$, i.e.~${\bf W}_{\kappa}$ is applied symmetrically to the points ${\bf x}_a$ and ${\bf x}_b$.
In that case, under the inverse transformation $\mathcal T^{-1} (.)$, we have $\delta_z=r\sin\phi_{ab}$ and $(-\delta_z)=r\sin\phi_{ba}$.
% $\delta_x=r\cos\phi_{ab}\cos\theta_{ab}$,
% $\delta_y=r\cos\phi_{ab}\sin\theta_{ab}$, $\delta_z=r\sin\phi_{ab}$, and
% $-\delta_x=r\cos\phi_{ba}\cos\theta_{ba}$,
% $-\delta_y=r\cos\phi_{ba}\sin\theta_{ba}$, $\delta_z=r\sin\phi_{ba}$.
The condition $\Phi_{k_\phi}\cdot\Phi_{k_\phi+1}\geq 0$ entails that $-\delta_z^2 = \delta_z\cdot(-\delta_z)=(r\sin\phi_{ab})\cdot(r\sin\phi_{ba})=r^2(\sin\phi_{ab}\sin\phi_{ba})\geq 0\Longrightarrow\delta_z=0$. Similarly,
$\Theta_{k_\theta}\cdot\Theta_{k_\theta+1}\geq 0 \Longrightarrow \delta_y=0$.
Since $\bold{x}_a\neq\bold{x}_b$, for  $\delta_x\neq0$ we have $ \cos\theta_{ab} = -\cos\theta_{ba} \Longrightarrow |\theta_{ab}-\theta_{ba}|=\pi$.
However, if  $\theta_{ab}$, $\theta_{ba}$ fall into the same bin, we have $|\theta_{ab}-\theta_{ba}|=\frac{2\pi}{n}<\pi$, which entails $\delta_x = 0$.  Thus, the assertion can not hold, and ${\bf W}_{\kappa}$ can not be applied to any two points symmetrically  unless both points are the same.

The asymmetry property of the spherical  kernel is significant because it restricts the sharing of the same weights between point pairs, which facilitates in learning more effective features with finer geometric details.
Lemma~2.1 also provides guidelines for the division of the convolution kernel into bins such that the asymmetry is always preserved.
%{\color{blue} In Fig. \ref{LemmaExamples}, we provide several examples of divisions that do not satisfy the asymmetry because of violating certain conditions of Lemma~2.1.}
%To elaborate further on this aspect, we provide few examples of kernel divisions that violate asymmetry in the supplementary material of the paper. 
Note that asymmetric application of kernel weights to pixels comes naturally in standard CNN kernels. However, the proposed kernel is able to ensure the same property in the point cloud domain.
\vspace{1mm}
%It is emphasize that preservation of asymmetric application of kernel weights to the points is  standard CNN kernels are also applied to pixel pairs asymmetrically. 

\begin{comment}
\begin{figure}
  \centering
   {\includegraphics[height=27.5mm]{lemma_examples.png}\label{LemmaExamples}}
  \caption{{\color{blue}Examples of divisions that violate the asymmetry. Here the red cross ${\bf x}_a$ represents the target point, while the green cross ${\bf x}_b$ is one of its neighborhood. The magenta sphere represents the spherical neighborhood range of ${\bf x}_a$, and the black curves represent divisions along $\theta$/$\phi$ directions.
  (a) The spherical space is not divided at all, which results in a single weight matrix to be defined in the kernel and applied to any two points ${\bf x}_a$ and ${\bf x}_b$ symmetrically. (b) 
  Let $\Phi=[-\frac{\pi}{2},\frac{\pi}{2}]$. There will be no divisions along the $\phi$ direction, which results in a particular weight matrix to be symmetrically applied to 
  points ${\bf x}_a$ and ${\bf x}_b$ on the $z$-$axis$ or its parallels (the blue arrow line). (c) 
  Let $\Theta=[-\pi,0,\pi]$, that is, $n=2$. The $\theta$ direction will be divided into two bins, which subtly results in   
%   the weight matrix in a certain bin with $\theta\in[0,\pi]$ to be symmetrically applied to 
%   points ${\bf x}_a$ and ${\bf x}_b$ on $x$-$axis$ or its parallel.
  points ${\bf x}_a$, ${\bf x}_b$ on the $x$-$axis$ and its parallels to share weights in a certain bin with $\theta\in[0,\pi]$ symmetrically.}} 
\end{figure}
\end{comment}
\begin{figure*}
  \centering
 {\includegraphics[height=40mm]{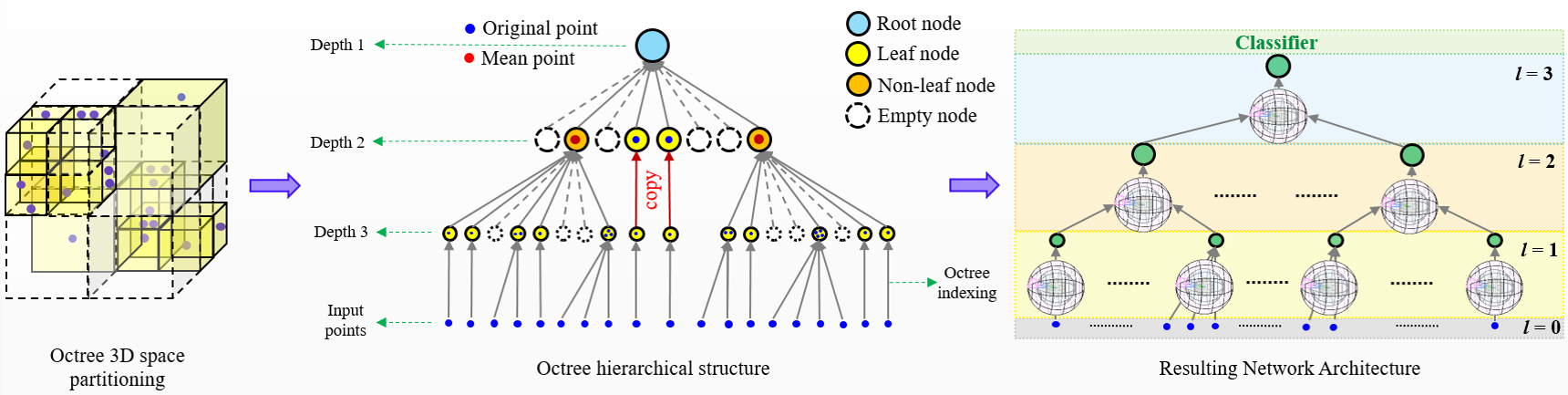}\label{octree}}
  \caption{Illustration of octree guided network architecture using a toy example: The point cloud in 3D space  is partitioned under an octree of depth 3. The corresponding tree representation allocates points to nodes at the maximum depth based on the  space partitioning, and computes the location of each parent node as the Expected location of its children. Leaf nodes on shallow branches are replicated to match the maximum depth.  The corresponding neural network has the same number of hidden layers as tree depth, and it learns spherical convolutional kernels for feature extraction.}
%  {\color{red}In the network architecture, I think it's better to add dots between the green nodes as well. Place P over cube. }}
  \label{fig:network}
  \vspace{-3mm}
\end{figure*}
\vspace{-3mm}
\noindent \textbf{Relation to 3D-CNN:}
Here, we briefly relate the proposed notion of spherical kernel  to the existing techniques that exploit CNNs for 3D data. Pioneering works in this direction  rasterize %transmute
the raw  data into uniform voxel grids, and then
extract features using 3D-CNNs from the resulting volumetric representations \cite{maturana2015voxnet,wu20153d}. In 3D-CNNs, the convolution kernel of size $3\times3\times3=27$ is prevalently used, that  splits the space in 1 cell/voxel for radius $r=0$ (self-convolution); 6 cells for radius $r=1$;
  12 cells for radius $r=\sqrt{2}$; and 8 cells for radius $r=\sqrt{3}$.
An analogous spherical convolution kernel for the same region can be specified with a radius $\rho=\sqrt{3}$, using the following edges for the bins:
 \begin{align}\label{split-sphConv}
 \notag
 &\boldsymbol{\Theta}=[-\pi,-\frac{\pi}{2},0,\frac{\pi}{2},\pi];\\
 \notag
&\boldsymbol{\Phi} =[-\frac{\pi}{2},-\frac{\pi}{4},0,\frac{\pi}{4},\frac{\pi}{2}];\\
&\bold{R} = [\epsilon,1,\sqrt{2},\rho], \epsilon\rightarrow0^+.
  \end{align}
  This division results in a \emph{kernel size} (i.e.~total number of bins) $4\times 4\times 3 + 1=49$, which is the coarsest multi-scale quantization allowed by Lemma~2.1. %\footnote{Further coarser quantization is allowed only under $\rho<\sqrt{3}$.}.

%   \footnote{Due to the importance of distance in 3D research, for kernels of identical size, we always prefer multi-scale quantization to single-scale quantization}.
  Notice that, if we move radially from the center to the periphery of  spherical kernel, we encounter identical number of bins (16 in this case) after each edge defined by  $\bold{R}$, where fine-grained bins are located close to the origin that can encode detailed local geometric information of the points. This is in sharp contrast to 3D-kernels that  must keep the size of all cells constant and rely on increased input resolution of the data to capture  finer details - generally entailing memory issues.
The multi-scale granularity of spherical kernel makes it a natural choice for raw point clouds. %Moreover, similar to 3D-CNN kernels, the spherical kernel also preserves the properties of translation-invariance and asymmetry.
\vspace{-1mm}
\section{Neural Network}
\vspace{-2mm}
% We use octree to build neurons of each layer and construct their neighborhoods for spatial convolution. The neuron construction is similar to the effect of voxelgrid sampling. And the tree structure avoids time-consuming knn/range searching. Although
% OctNet previously exploits octree to avoid dense voxelization of the 3D space, the kernel and raw features explored determine that we are completely different from it.
Most of the existing attempts to  process point clouds with neural networks  \cite{li2018so,li2018pointcnn,qi2017pointnetplusplus,shen2018mining,simonovsky2017dynamic}  rely on K-NN  or range searches to define local neighborhood of points, that are  subsequently used  to perform  operations like convolution or pooling. %Recent years have seen few attempts to directly process point clouds with neural networks,  e.g. PointNet++ \cite{qi2017pointnetplusplus}, ECC \cite{simonovsky2017dynamic}. These works predominately rely on K-NN  or range searches to build local regions around the points to perform operations like convolution. 
However, to process large point clouds, these search strategies become computationally prohibitive.   For unstructured data, an efficient mechanism to define point neighbourhood  is tree-structuring, e.g.~Kd-tree~\cite{bentley1975multidimensional}.  The hierarchical nature of tree structures also provide guidelines for  neural network architectures that can be used to process the point cloud. More importantly, a tree-structured data also possess
the much desired attributes of permutation and translation invariance for  neural networks.

\subsection{Core Architecture}
\vspace{-2mm}
We exploit octree structuring \cite{meagher1982geometric}  of point clouds and design a  neural network  based on the resulting trees. Our choice of using octree comes from its   amenability to neural networks as the base data structure~\cite{riegler2017octnet},  and its ability to account for more data in point neighborhoods compared to, for example, Kd-tree.  We illustrate 3D space partitioning  under octree, the resulting tree, and the formation of neural network using the proposed strategy of network construction  in Fig.~\ref{fig:network} for a toy example. For an input point cloud $\mathcal{P}$, we  construct an octree of depth $L$  ($L = 3$ in the figure). In the  construction, the splitting of  nodes is fixed to use a maximum  capacity of one point, with the exception of the last layer leaf nodes. The point in a parent node is computed as the Expected value of the points in its children. The allocation of multiple points in the last layer  nodes directly results from the allowed finest partitioning of the space. For the sub-volumes in 3D space that are not densely populated, our splitting strategy can result in leaf nodes before the tree reaches its maximum depth. In such cases, to facilitate mapping of the tree  to  a neural network, we  replicate the leaf nodes to the maximum depth of the tree.
We safely ignore the empty  nodes while implementing the network,  resulting in computational and memory benefits.

Based on the hierarchical tree structure, our neural network also has $L$ hidden layers. Notice that, in Fig.~\ref{fig:network}  we use $l = 1$ for the first hidden layer that  corresponds to Depth $= 3$ for the tree. We will use the same convention in the text to follow. For each non-empty node in the tree, there is a corresponding neuron in our neural network.  Recall that, a spherical convolutional kernel is specified with a target point over whose neighborhood the convolution is performed. Therefore, to facilitate convolutions, we associate a single 3D point  with each neuron, %These locations are computed as the Expected values of the children of each node in the octree,
except for the leaf nodes at the maximum depth of the tree. For a leaf node, the associated point is the mean value of data points allocated to that node.
A neuron uses its associated point/location to select the appropriate spherical kernel and later applies the non-linear activation (not shown in Fig.~\ref{fig:network}).  In our network, all  convolution layers before the last layer are followed by batch normalization 
%\footnote{The number of neurons in the same layer/level/depth of the octree may vary across different samples. We conduct an unconventional batch normalzation by forcing different neurons to share the same expectations and variances. The batch normaliztion contribute a lot to the training speed.} 
and ReLU activations.

We denote the location associated with the $i^{\text{th}}$ neuron in the $l^{\text{th}}$ layer of the network as $\bold{\bar{x}}_i^l$. From $l =1$ to $l = L$, we can  represent the locations associated with all neurons as $\mathcal{Q}^1=\{\bold{\bar{x}}_i^1\}_{i=1}^{m_1}$, $\dots$, $\mathcal{Q}^L=\{\bold{\bar{x}}_1^L\}_{i = 1}^{m_L}$. Denoting the raw input points as $\mathcal{Q}^0=\{\bold{\bar{x}}_i^0\}_{i=1}^{m_0}$, $\bold{\bar{x}}_i^l$ is numerically computed by our network   as:
\vspace{-3mm}
\begin{align}\label{node-xyz}
  \bold{\bar{x}}^l_i = \frac{\sum\limits_{\bold{\bar{x}}_j^{l-1}\in \mathcal{N}(\bold{\bar{x}}^l_i)}\bold{\bar{x}}_j^{l-1}}{|\mathcal{N}(\bold{\bar{x}}^l_i)|},
  \end{align}
  where $\mathcal{N}(\bold{\bar{x}}_i^l)$ contains  locations of the relevant children nodes in the octree.  It is worth noting that the strategy used  for specifying the network layers also entails that $|\mathcal Q^{l-1}| > |\mathcal Q^l|$.  %to the end that there is only one location associated with the final hidden layer corresponding to the root of the octree.
Thus, from the first layer to the last, the features learned by our network move from lower to higher level of abstraction similar to the standard CNNs.

\begin{figure*}[t]
  \centering
  \includegraphics[width=\textwidth]{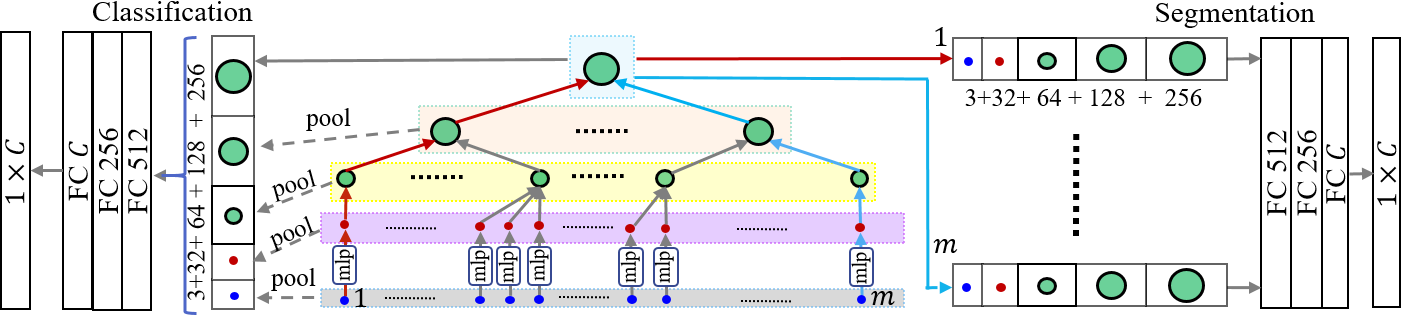}
  \caption{Classification and segmentation using the core network of Fig.~\ref{fig:network}. For classification, the features at the root node (top layer) are concatenated with the max-pooled (dashed lines) features at the remaining layers followed by FC layers. For segmentation, the representation of a point uses the layer-level features of all the ancestors along the path to the root node, e.g.~red path for point `1' and blue path for point `$m$'.  Point-wise classification (segmentation) is performed using the concatenated raw point features ($xyz/xyz-rgb$), the MLP features and all the extracted layer-level features. A simple configuration MLP(32)-Octree(64-128-256) is shown for illustration.}
  %\label{sphConv}
  \label{fig:segNet}
  \vspace{-3mm}
\end{figure*}

In relating the spherical nature of point  neighborhood  considered in our network to  the cubic partitioning of  space by octree, a subtle detail is worth considering.
Say $\bold{x}_{\min}$ = $[x_{\min}, y_{\min}, z_{\min}]^{\intercal}$, and $\bold{x}_{\max}$ = $[x_{\max}, y_{\max}, z_{\max}]^{\intercal}$  determine the range of point coordinates in a given cubic volume resulting from our space partitioning.  The spherical neighborhood associated  with a  neuron in the $l^{\text{th}}$ layer is defined with  the radius $\rho=2^{l-L-1} || \bold{x}_{\max} -\bold{x}_{\min} ||_2$.  %\cdot d(\bold{x}_{\min},\bold{x}_{\max})$, 
This neighbourhood may not strictly circumscribe all points of the corresponding cubic volume at this level due to shape dissimilarity. Although  the number of such points is minuscule in practice,  we still take those  into account by assigning them to the outer-most bins of our kernels based on their azimuth and elevation values.

Our neural network performs inter-layer convolutions instead of intra-layer convolutions. This drastically reduces the operations required to process large point clouds when compared with graph-based networks  \cite{bruna2013spectral,defferrard2016convolutional,kipf2017semi,simonovsky2017dynamic,yi2017syncspeccnn}. We note that for all nodes with a single child, only self-convolutions are performed in the network. Note that due to its unconventional nature, spherical convolutional kernel  is not readily implemented using the existing deep learning libraries, e.g.~matconvnet~\cite{vedaldi2015matconvnet}. Therefore, we implement it ourselves with CUDA C++ and mex interface\footnote{The implementation will be made public.}. For the other modules such as ReLU, batch normalization etc., we use matconvnet.

\vspace{-5mm}
{\paragraph{Comparison to OctNet~\cite{riegler2017octnet}:}   OctNet~\cite{riegler2017octnet}  also makes use of octree structure. However, OctNet processes point clouds as regular 3D volumes - a  3D-CNN.
In  contrast, we process point clouds following their  unstructured nature.
Our network  learns features  for each point in the sets from $\mathcal{Q}^0$ to $\mathcal{Q}^L$, which is in contrast to OctNet that must account for occupied and unoccupied voxels, entailing complexity.  
We exploit octree structure to simultaneously construct  neighborhoods of all points and coarsen the original point cloud layer-by-layer, while OctNet uses this structure to voxelize the point cloud into different resolutions.
% Moreover,  differnt from OctNet, octree structuring in our network is also utilized  for 
%coarsening the original point cloud in a  layer-by-layer manner. 

%Despite the same exploration of octree structure, OctNet \cite{riegler2016octnet} is essentially
%a 3D-CNN architecture which processes point clouds as regular 3D volumes.
%To learn features, it has to account for both the occupied and unoccupied voxels. In sharp contrast, we process the point cloud following its irregular nature, and learn features only for each point in the sets from $\mathcal{Q}^0$ to $\mathcal{Q}^L$. In our work, octree is used to
%coarsen the original point cloud layer by layer, and meanwhile construct the neighborhoods of all points efficiently.}
\subsection{Classification and Segmentation}
\vspace{-2mm}
The classification and segmentation networks are basically variants of the same core architecture shown in Fig~\ref{fig:network}. 
However, we additionally insert an MLP layer prior to the octree structure to obtain more expressive point-wise features. This concept is inspired from Kd-Net~\cite{klokov2017escape}. Figure \ref{fig:segNet} shows the complete architectures for classification and segmentation. To fully exploit the hierarchical features learned at different octree levels, we use features from all octree layers. For classification, we max pool the features from intermediate layers, including the raw features, and concatenate them with the features at the root node to form a global representation of the complete point cloud. For segmentation, we need point wise features. The feature of each point is the concatenation of raw features, MLP features and layer-wise features without any pooling. The final classification or segmentation is performed using three fully connected layers.

\section{Experiments}
\vspace{-2mm}
%{\color{blue} 
We conduct experiments on clean CAD Models as well as noisy  point clouds to evaluate the performance of our method for the tasks of 3D object classification, part segmentation and semantic segmentation. Throughout the experiments, we keep the size of our convolution kernel fixed to  $8\times2\times3+1$, in which the radial dimension is split uniformly. We use three fully connected layers (512-256-$C$) followed by softmax as the classifier for both the classification and segmentation tasks. Here, $C$ denotes the number of classes/parts.
The training of our network is conducted using a Titan Xp GPU with 12 GB memory.
We use Stochastic Gradient Descent with momentum to train the network. The batch size  is kept fixed to 16 in all our experiments.
These hyper-parameters were empirically optimized using cross-validation.
We use only the  $(x,y,z)$ coordinates of points provided by point clouds to train our network, and the $(r,g,b)$ values when the color information is provided.
Few existing methods in the literature also take advantage of normals, and use them as input features. 
However, normals are not directly sensed by 3D sensors and must be computed using the point coordinates. This also entails additional computational burden. Hence, we avoid using normals as input features.
%In addition, there are instances in literature where normals are also exploited as input features. However, normals may not always be readily available for point clouds, which then results in undesired computational overhead to be obtained. In this work, we restrict the input features of our network to be $(x,y,z)$ coordinates as well as $(r,g,b)$ values when the color information is provided.}
%{\color{blue}\paragraph{Data augmentation:} 
%\vspace{-2mm}
In our experiments, we follow the standard practice of taking advantage of data augmentation. 
To that end, we used random sub-sampling of the original  point clouds, performed random azimuth rotation (up to $\frac{\pi}{6}$ rad) and also applied noisy translation (std.~dev = 0.02) to increase the number of training examples. These operations were performed on the fly in each training epoch of the network.  %Similar augmentations are applied to the test examples in the test stage.
%}
\subsection{Classification}
\vspace{-2mm}
We use the  benchmark datasets ModelNet10 and ModelNet40 ~\cite{wu20153d} to evaluate our technique for the classification task.
These  datasets are created using clean CAD models.  ModelNet10 contains 10 categories of object meshes, and the samples are split into 3,991 training examples and 908 test instances. ModelNet40 contains object meshes for 40 categories with 9,843/2,468 training/testing split.

Compared to existing works  
(e.g.~\cite{qi2017pointnet,qi2017pointnetplusplus,shen2018mining,simonovsky2017dynamic}), the  convolutions performed in our network allow the proposed method to consume large input point clouds. Hence, we train our network using 10K input points. For the classification task, we adopted a network with 6 levels of octree, whereas the number of feature channels are kept MLP(32)-Octree(64-64-64-128-128-128). The network comprises two components, octree based architecture for feature extraction and classification stage. We train the whole network in an end-to-end fashion. We  standardize the input models by normalizing the 3D point
clouds to fit into a cube of $[-1,1]^3$ with zero mean.

Table~\ref{compare2Others} benchmarks the object classification performance of our approach that is abbreviated as $\Psi$-CNN\footnote{A Greek alphabet is chosen as prefix to avoid duplication with other OCNNs and SCNNs, e.g. \cite{liu2015sparse,parashar2017scnn,wang2017cnn}.}.  Our method uses $xyz$ coordinates of points as raw features to achieve these results.
%Among the reported `class' and `instance' accuracies, the former is the average accuracy per object category, whereas the latter percentage of correctly classified instances. 
%{\color{blue}We abbreviate our approach as $\Psi$-CNN\footnote{A Greek alphabet is chosen as prefix to avoid duplication with other SCNNs, e.g. \cite{liu2015sparse,parashar2017scnn}.} in
%Table~\ref{compare2Others} that summarizes the object classification performance on the ModelNets with $xyz$ coordinates as raw features.
%Along the `class' and `instance' accuracies, the former is the average accuracy per object category.
As can be seen, $\Psi$-CNN consistently achieves the best performance on ModelNets. 
%on-par or better results than the latest methods, SO-Net \cite{li2018so} and KCNet \cite{shen2018mining}, for point cloud classification. 
We note that, like our method Kd-Net \cite{klokov2017escape} and OctNet \cite{riegler2017octnet} are also tree structure based networks. However, they require twice the number of parametric layers as required by our method to achieve the reported performance. 
This is a direct consequence of effective exploration of geometric information by the proposed kernel. %We also provide an ablation study to support this in the supplementary material of the paper. 

\begin{table}[t]
\centering
\caption{Classification performance on ModelNets \cite{wu20153d}.}\label{compare2Others}
\begin{tabular}{l|c|c|c|c}
  \hline
  \multirow{ 2}{*}{Method}&  \multicolumn{2}{c|}{ModelNet10} & \multicolumn{2}{c}{ModelNet40} \\
  \cline{2-5}
  %\hline
%   \multicolumn{3}{l|}{Classes} & \multicolumn{2}{c|}{10} & \multicolumn{2}{c}{40} \\
%   \hline
 &class & instance & class & instance \\
  \hline
  \hline
  OctNet~\cite{riegler2017octnet} &90.1 & 90.9 & 83.8 & 86.5 \\
  ECC~\cite{simonovsky2017dynamic}&90.0&90.8 & 83.2&87.4\\
  PointNet~\cite{qi2017pointnet} & -- & -- & 86.2 &89.2 \\
  PointNet++~\cite{qi2017pointnetplusplus}& -- & -- & --  &90.7 \\
%   PointNet++&$xyz$+normal&9 & -- & -- & --  &91.9 \\
  %Kd-network (DT) &$xyz$&11 & -- &89.2 &-- &85.7\\
  Kd-Net~\cite{klokov2017escape} & 92.8 &93.3 &86.3 &90.6\\
%    Kd-network &$xyz$&16 & 93.5&94.0& 88.5&91.8\\
  SO-Net~\cite{li2018so}&93.9 &94.1 &87.3 &90.9 \\
  KCNet~\cite{shen2018mining} &--& 94.4 &--& 91.0\\
  \hline
%    $\Psi$-CNN (proposed) & $xyz$ & 7 & \textbf{94.0} & \textbf{94.2} & \textbf{88.4}  & \textbf{92.0}  \\
$\Psi$-CNN &  \textbf{94.4} & \textbf{94.6} & \textbf{88.7}  & \textbf{92.0}  \\
% $\Psi$-CNN  &  \textbf{xx} & \textbf{xx} & \textbf{xx}  & \textbf{xx}  \\
  \hline
\end{tabular}
\vspace{-3mm}
\end{table}

%It is worth mentioning that among
%the compared approaches, only Kd-Net \cite{klokov2017escape} and OctNet \cite{riegler2016octnet} are based on tree structures, like $\Psi$-CNN. Yet, the two methods produce their results by using almost twice the number of parametric layers than ours. 
%Compared to them, we attribute our performance boost mainly to the accurate exploration of geometric information by spherical kernels for which we tailor our network. {\color{red}To support this argument further, we provide an ablation study in the supplementary material, which removes the max pooling and hierarchical concatenation, but uses the octree root-node feature only as the global representation of the input point cloud.}
%}
\subsection{Part Segmentation}
ShapeNet part segmentation dataset \cite{yi2016scalable} contains 16,881  CAD models from 16 categories. The models in each category have two to five annotated parts, amounting to 50 parts in total. The point clouds are created with uniform sampling from 3D meshes.
This dataset provides  $xyz$ coordinates of the points as raw features, and has 14007/2874 training/testing split defined. 
%{\color{red} Please refer to the supplementary material for the training details.} %We formulate part segmentation as a per-point classification problem. 
We  use a 6-level octree for the segmentation network, with configuration MLP(64)-Octree(128-128-256-256-512-512). The output class number $C$ of the classifier is determined by the number of parts in each category.
We use the part-averaged IoU (mIoU)  proposed in \cite{qi2017pointnet} to report the performance in Table \ref{table:partSeg}. Similar to the classification task, we also standardize the input models of ShapeNet by normalizing input point
clouds to  $[-1,1]^3$ cube with zero mean.

%{\color{blue}
In Table \ref{table:partSeg}, we compare our results with the popular methods that also take irregular point clouds as  input. Yet, to achieve their results, some of these methods exploit  normals besides $xyz$ coordinates as input features, e.g.~PointNet, PointNet++, SO-Net. 
It can seen that $\Psi$-CNN not only achieves the highest mIoU $86.8\%$, but also outperforms the other approaches on 11 out of 16 categories.
To the best of our knowledge, $\Psi$-CNN records the new state-of-the-art performance on this part segmentation dataset that is $\sim1\%$ higher than 
the specialized segmentation networks, SSCN~\cite{graham20183d} and SGPN \cite{wang2018sgpn}.

%{\color{red}We provide a thorough table in the supplementary material that compares our architecture with more existing works, no matter they take the raw point cloud as input or not.}

In Fig.~\ref{partseg_examples}, we show few representative segmentation results. High mIoU is achieved by $\Psi$-CNN for the high-quality results, whereas the mIoU value is low for the other case.  Examining the low-quality results, we found that most of these cases are caused by one of the two conditions. (1)~Confusing ground truth labelling: E.g.~the axle in \textit{Skateboard} is labelled as a separate  segment in most of the ground truth samples but part of the wheels in few other samples. Hence, the network learns the more dominant segmentation. Similar is the case for the legs of \textit{Chairs}. (2)~Small parts without clear boundaries: E.g.~handles of a \textit{Bag} are considered separate segments in the ground truth.  %We also provide further examples in the supplementary material. 
From these results, we can easily conclude the success of $\Psi$-CNN for the part segmentation task.

%In Fig.~\ref{partseg_examples}, we show examples of both high-quality and low-quality segmentations generated by our network. We notice that among the low-quality segmentations,
%some cases are caused because the parts are too delicate to be learned correctly, like the small handles of the \emph{bag} and the tiny lights of the \emph{lamp}. Other cases are mostly caused by the confusing labelling cast by the ground truth. For example, in the ground
%truth of \emph{skateboard}, the axles are sometimes labelled separately from the wheels and sometimes not. Similarly, in the \emph{Chair}
%category, the back and legs are sometimes labelled as separate parts and sometimes not. For networks which make use of $xyz$ coordinates information only, such ambiguous ground truth labelling confuses the network, and it tends to perform predictions based on the dominant cases in each category. 
%{\color{red} More examples are provided in the supplementary material.}
%} 

%In particular, this metric is computed as follows. 
%For 
%each shape S of category C, to calculate the shape's mIoU:
%For each part type in category C, compute IoU between
%groundtruth and prediction. If the union of groundtruth and
%prediction points is empty, then count part IoU as 1. Then
%we average IoUs for all part types in category C to get mIoU
%for that shape. To calculate mIoU for the category, we take
%average of mIoUs for all shapes in that category.
\begin{figure*}[!t]
  \centering
   \begin{tabular}{cc|cc!{\vrule width1.2pt}cc|cc}
   \hline
   \multicolumn{4}{c!{\vrule width1.2pt}}{
   High-Quality Segmentation} & \multicolumn{4}{c}{
   Low-Quality Segmentation}\\
   \hline
     GT & Ours & GT &Ours & GT &Ours & GT &Ours \\
     \hline
     %& & & & & & & &
     \includegraphics[width=16mm,height=17mm]{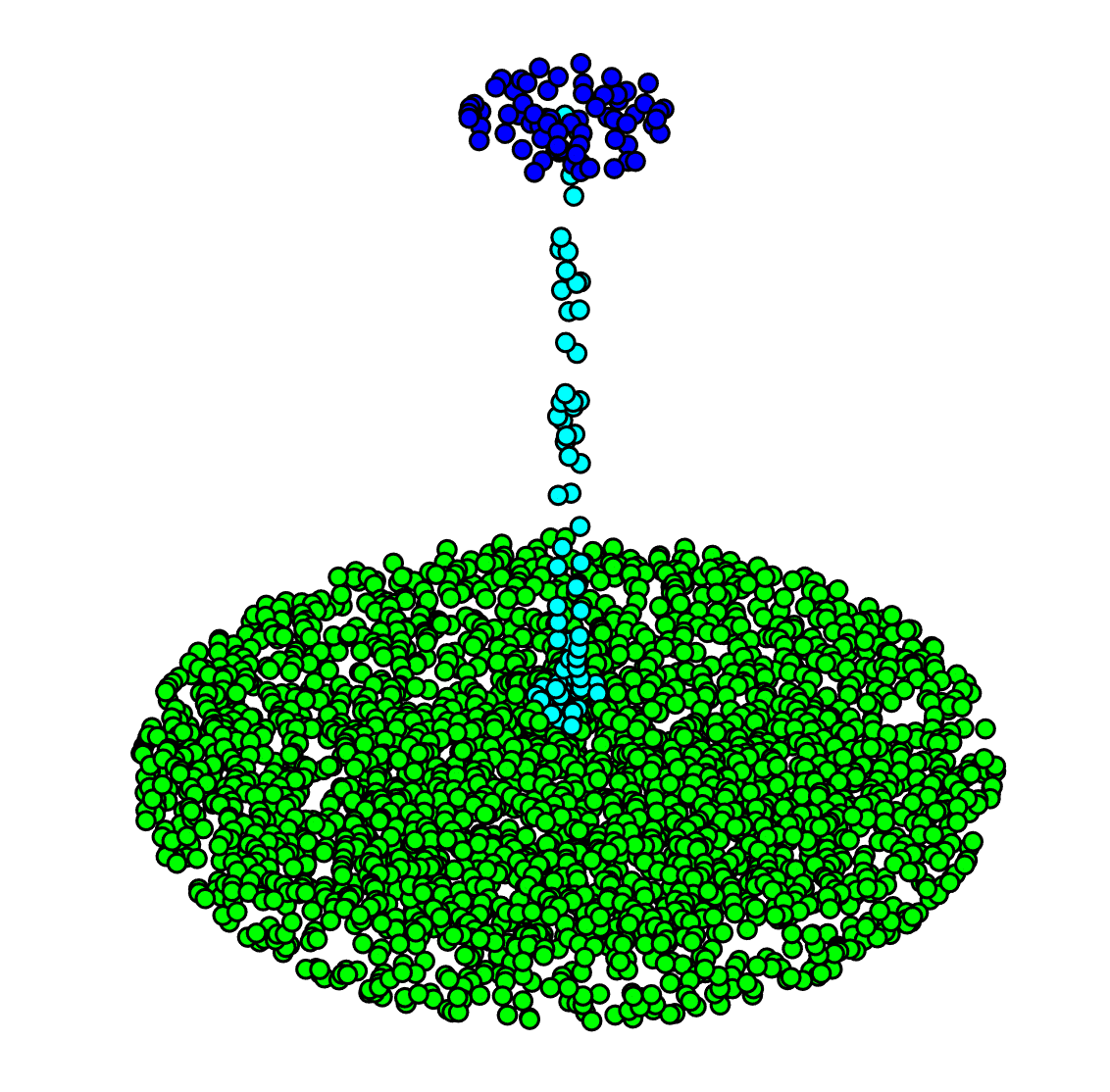} & 
     \includegraphics[width=16mm,height=17mm]{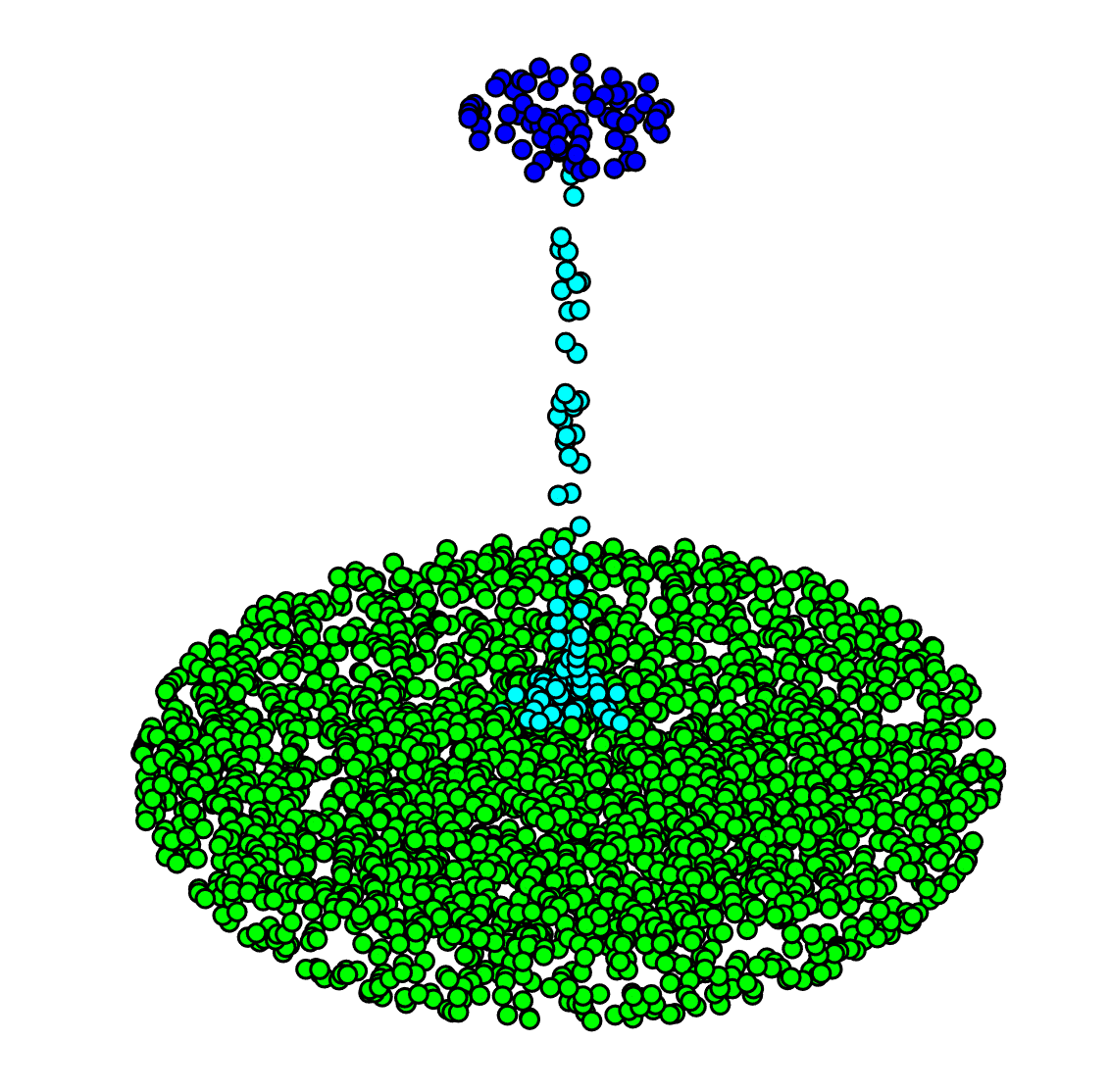} &
     \includegraphics[width=16mm,height=17mm]{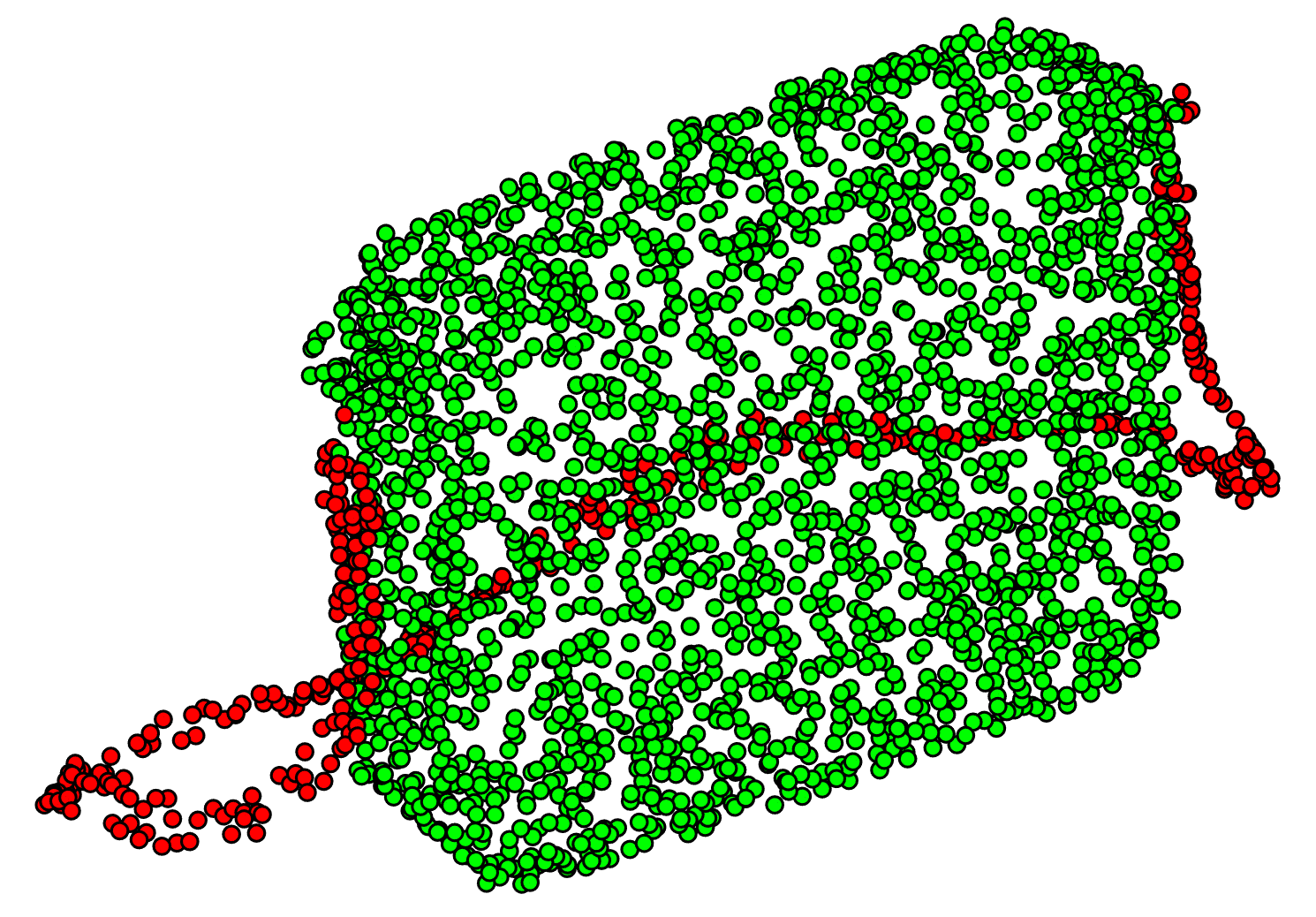} &
     \includegraphics[width=16mm,height=17mm]{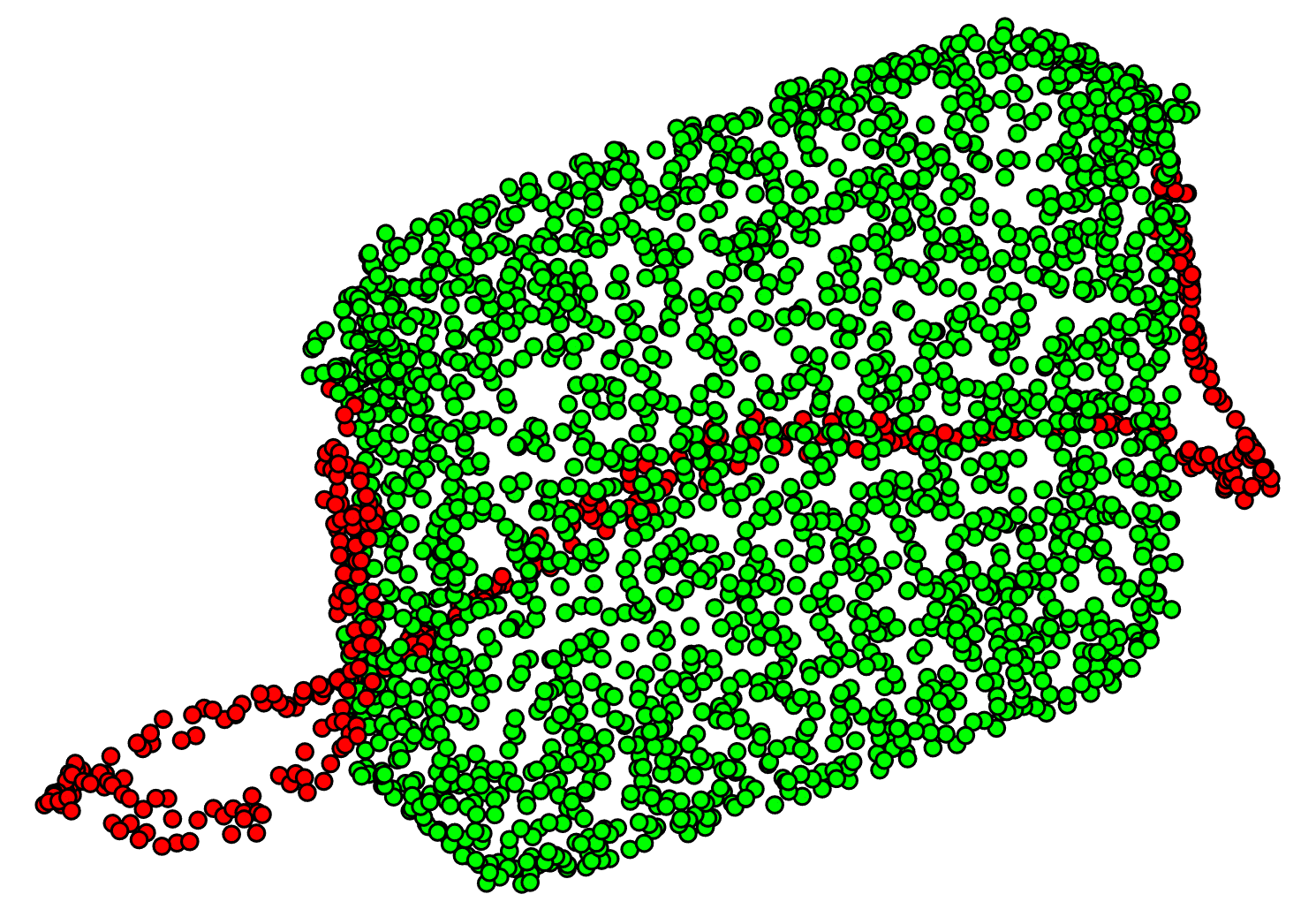}  &
   \includegraphics[width=16mm,height=17mm]{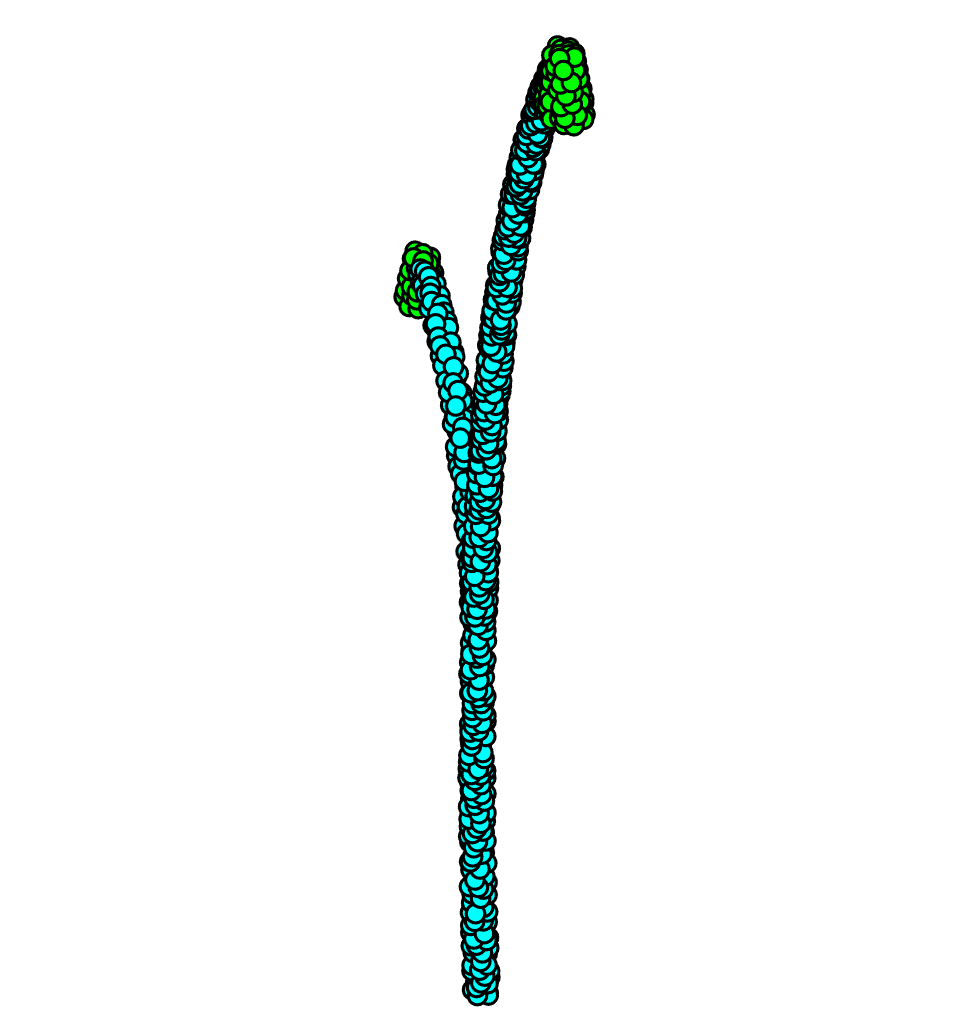} & 
   \includegraphics[width=16mm,height=17mm]{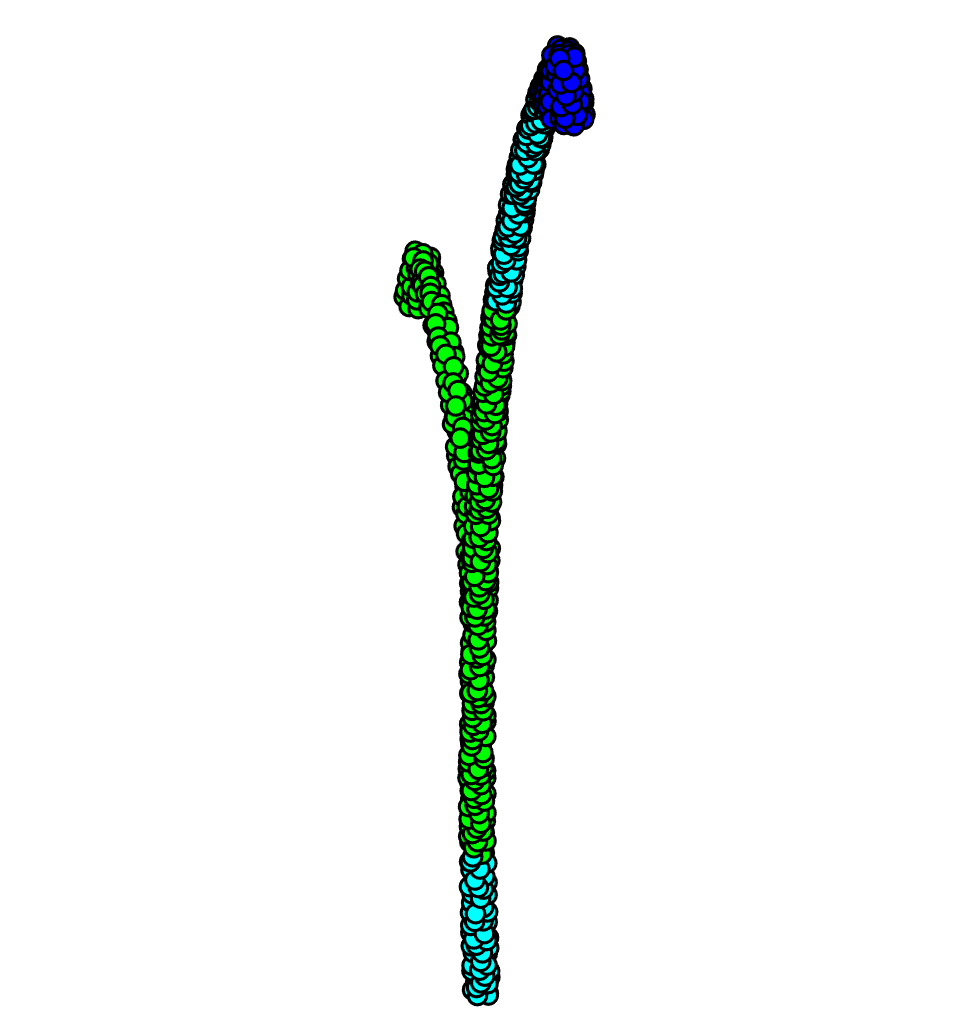} &
     \includegraphics[width=16mm,height=17mm]{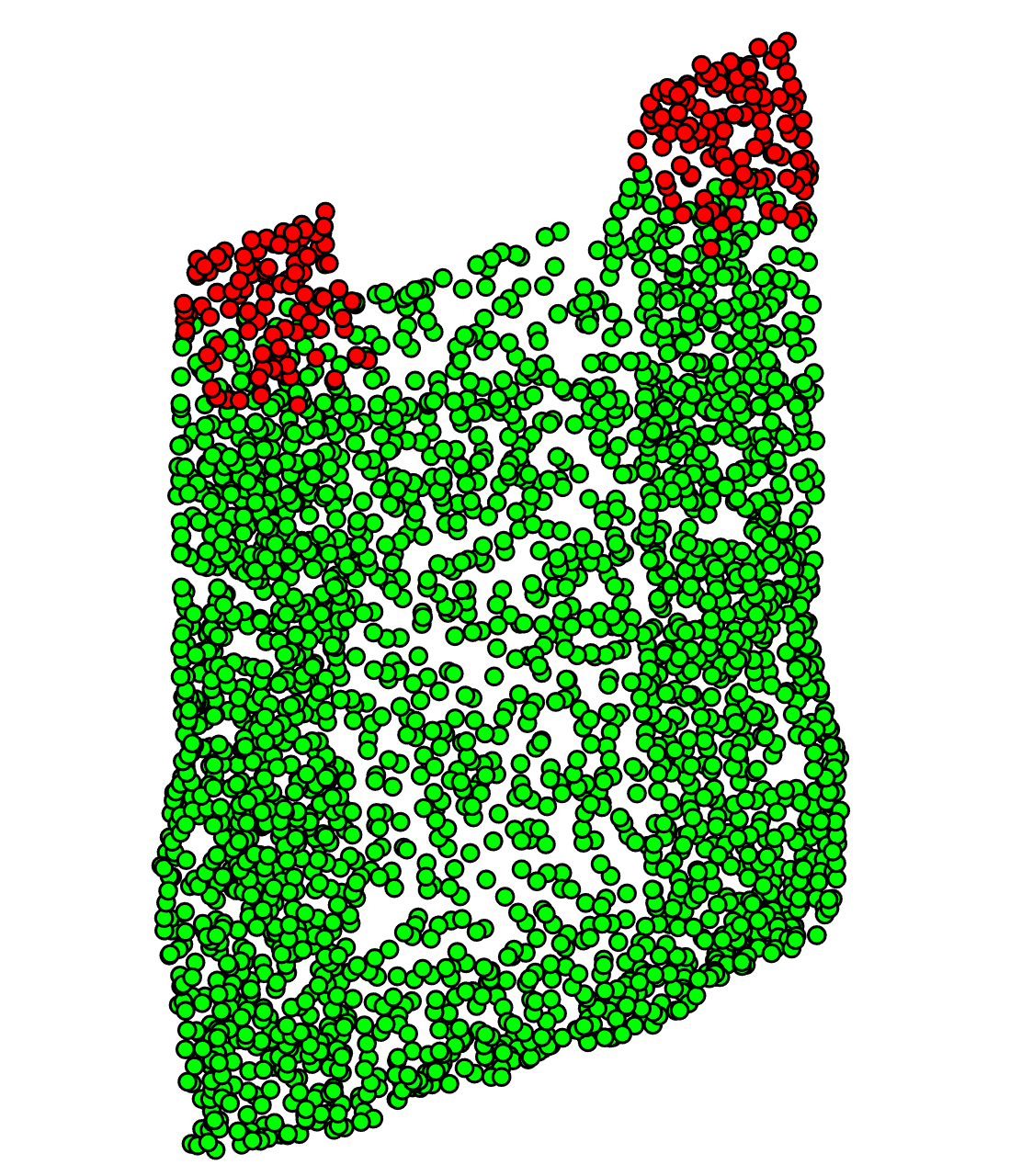} 
     & \includegraphics[width=16mm,height=17mm]{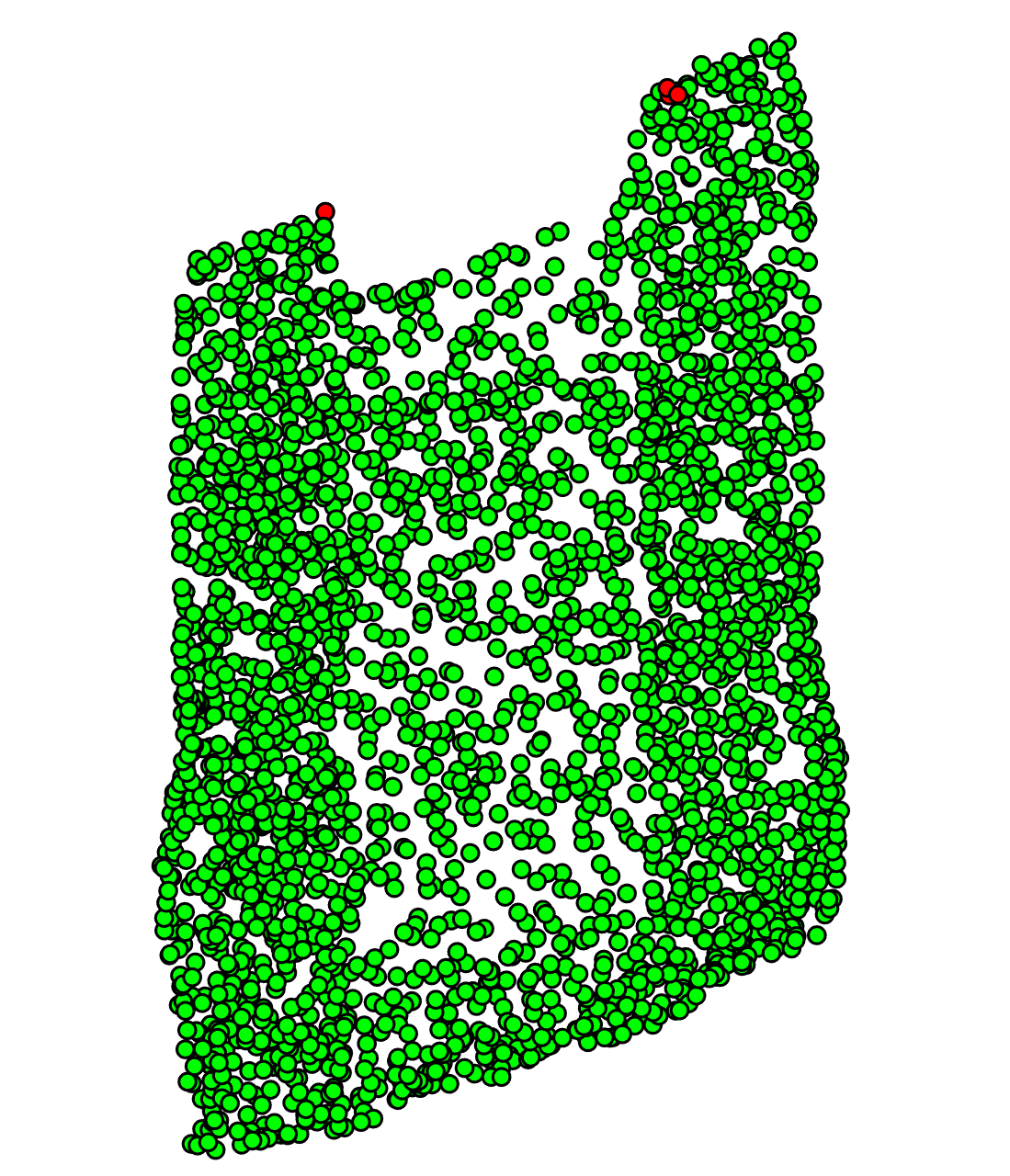}  \\
   Lamp & 91.4\% & Bag &98.1\% & Lamp  &35.5\% & Bag &46.8\% \\
   \hline
  % & & & & & & & &
    \includegraphics[width=16mm,height=17mm]{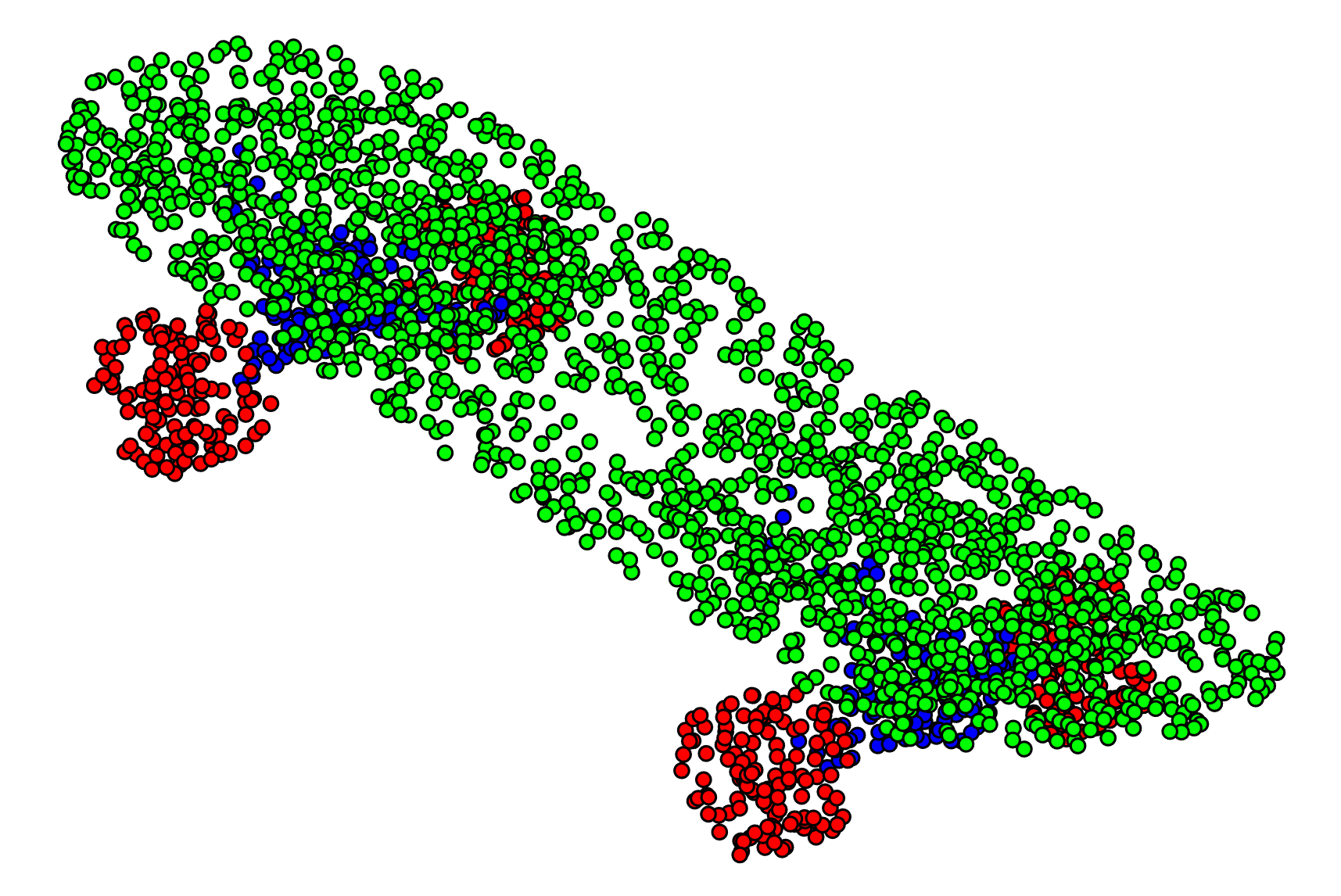} & 
    \includegraphics[width=16mm,height=17mm]{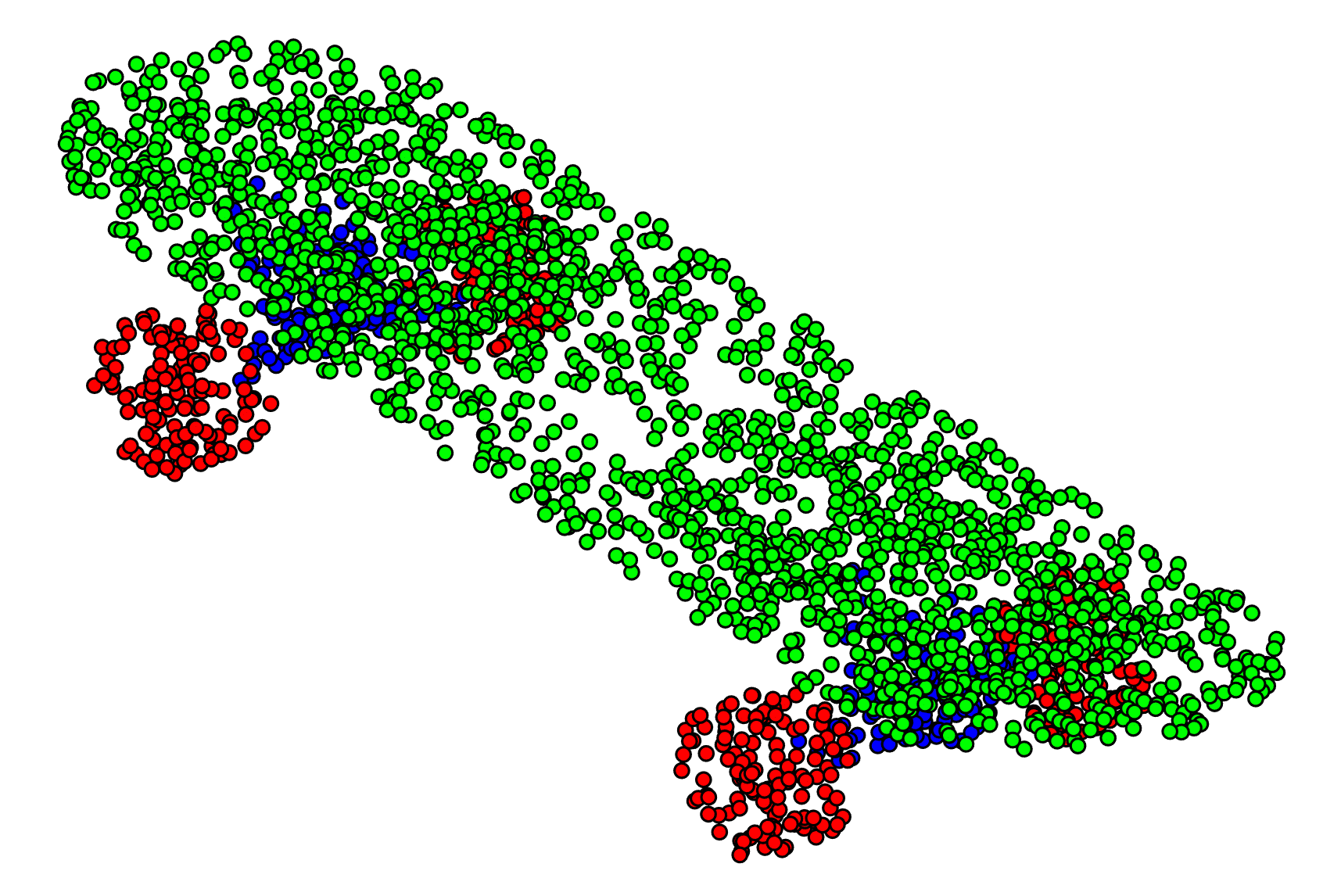} &
    \includegraphics[width=16mm,height=17mm]{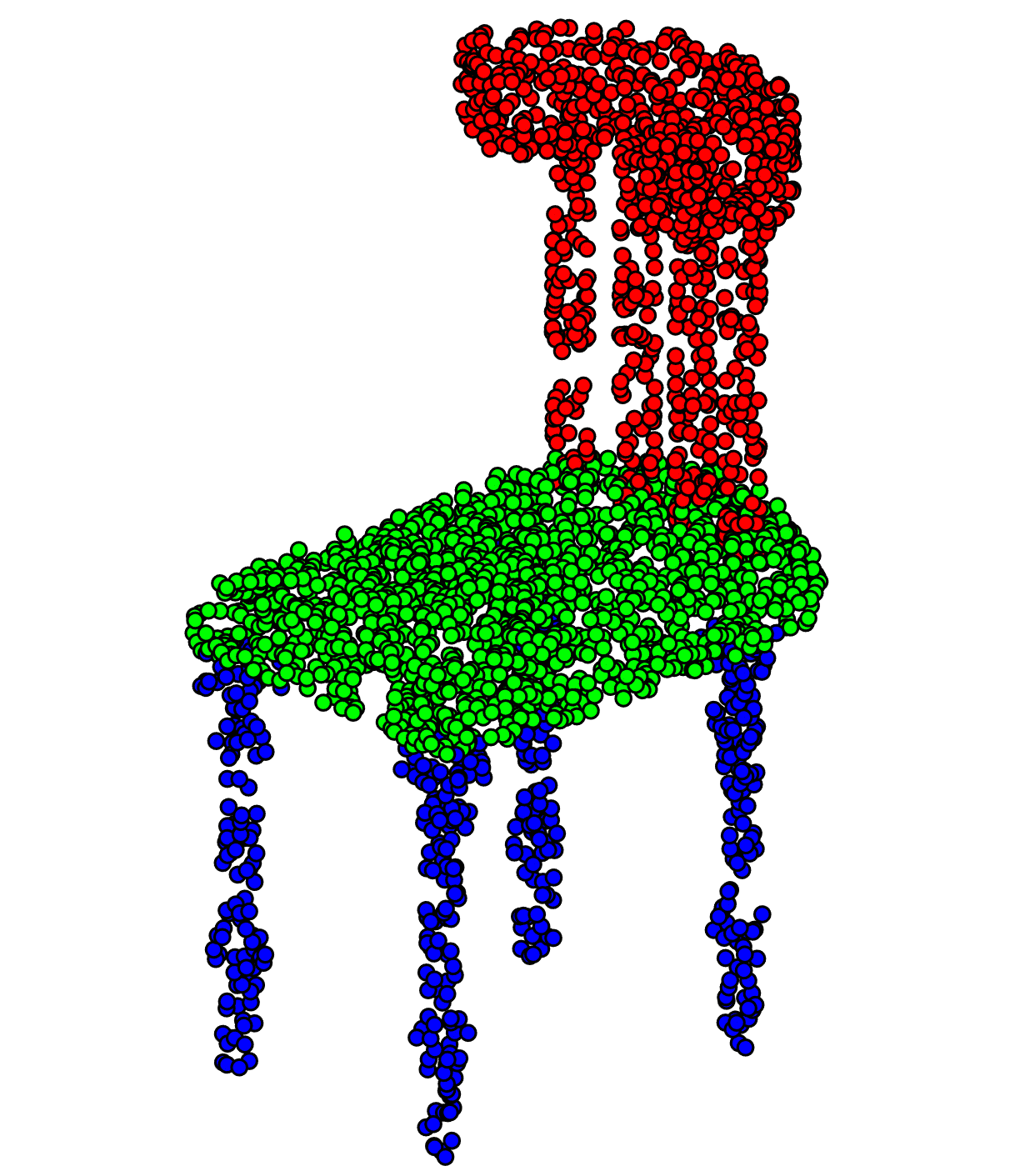}   & 
    \includegraphics[width=16mm,height=17mm]{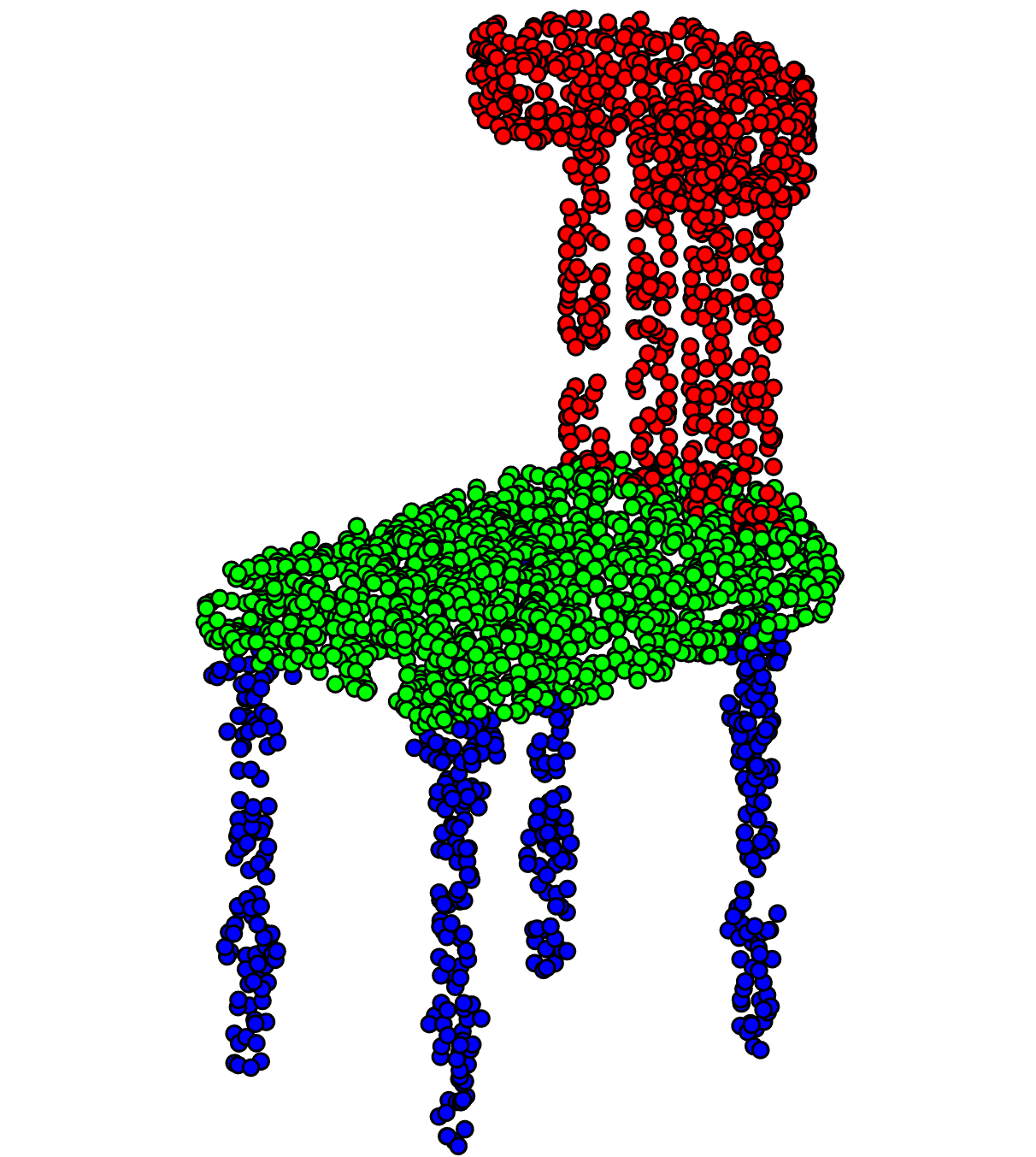} & 
    \includegraphics[width=16mm,height=17mm]{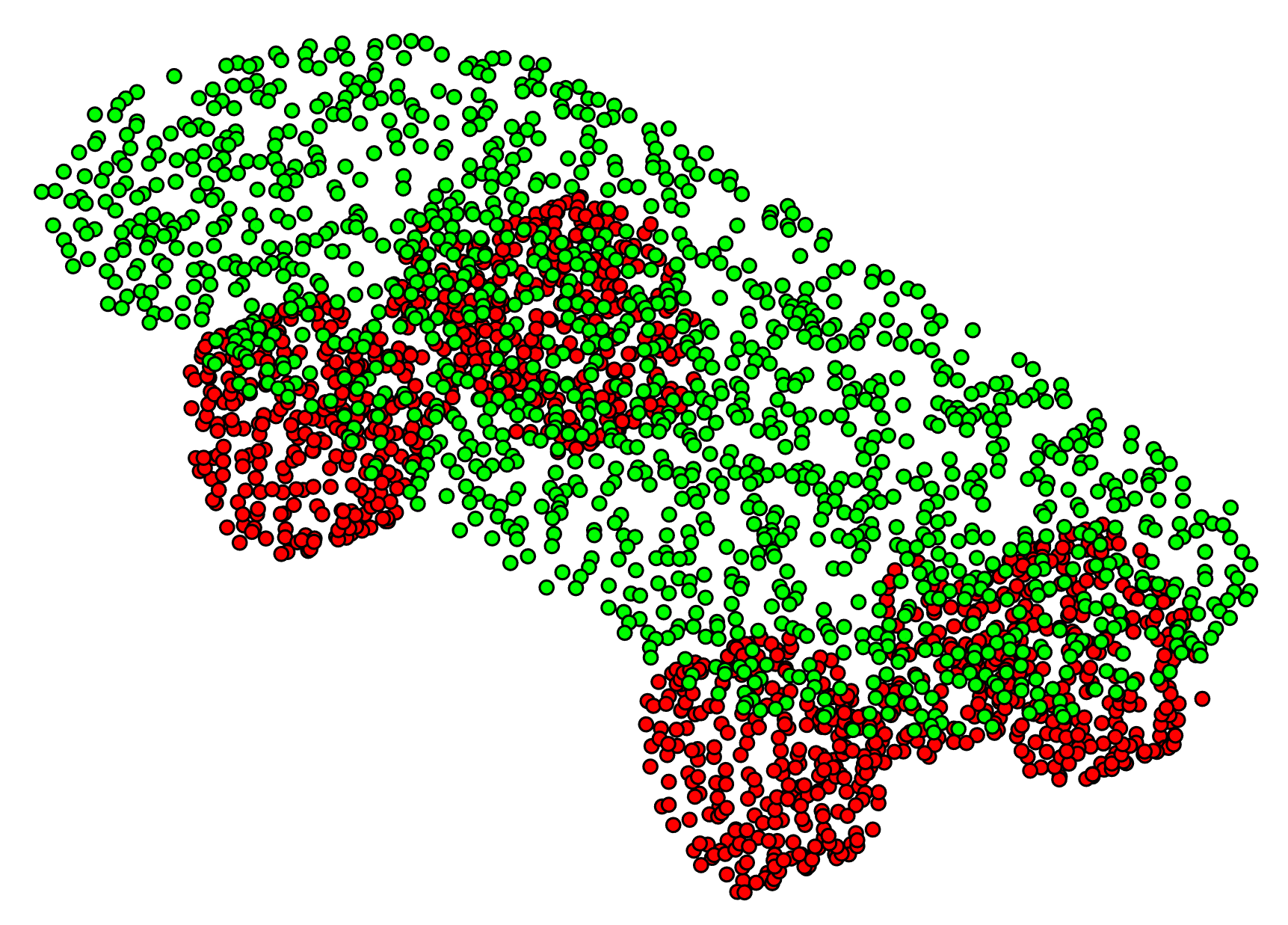} & 
    \includegraphics[width=16mm,height=17mm]{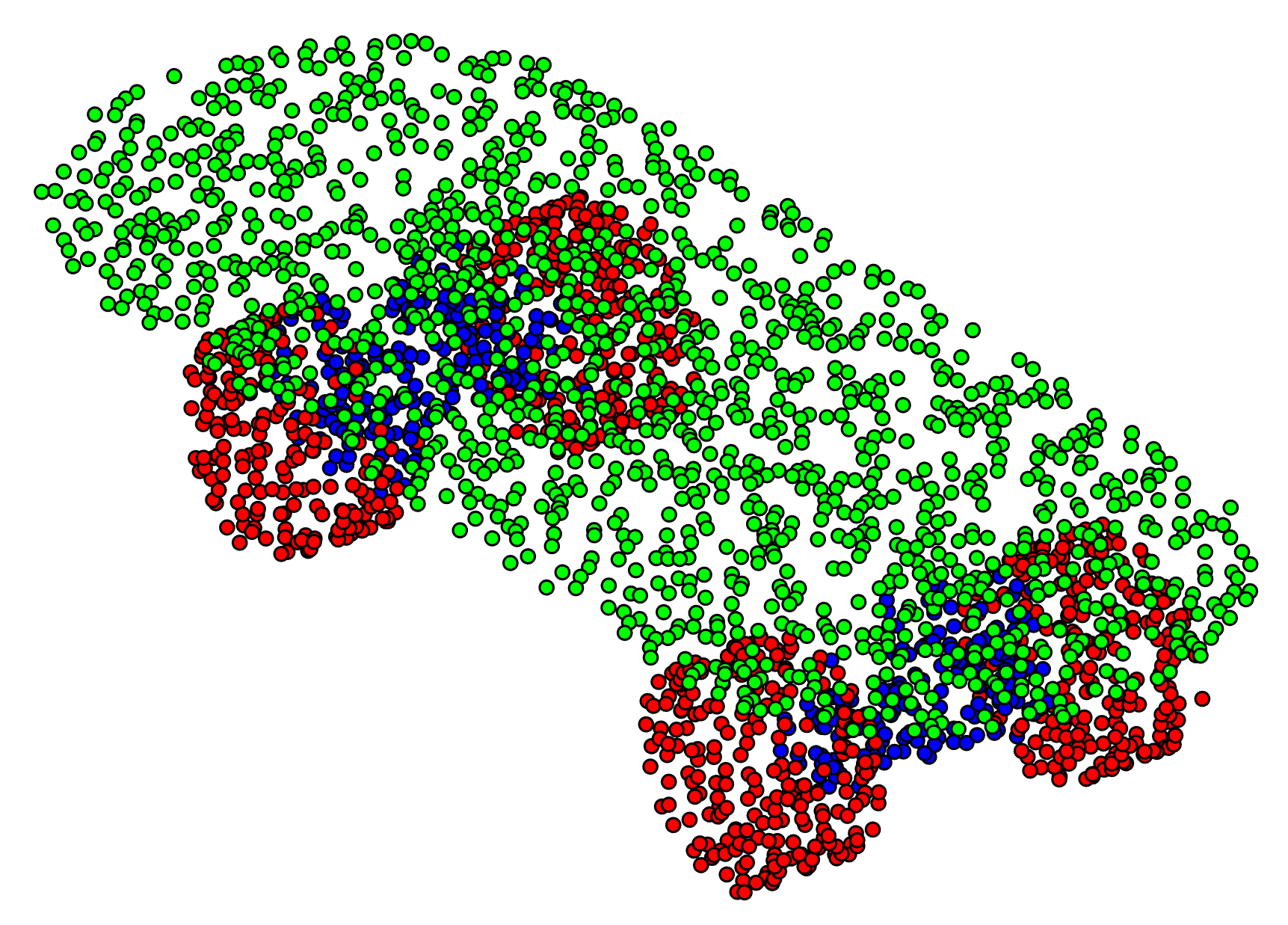} &
    \includegraphics[width=16mm,height=17mm]{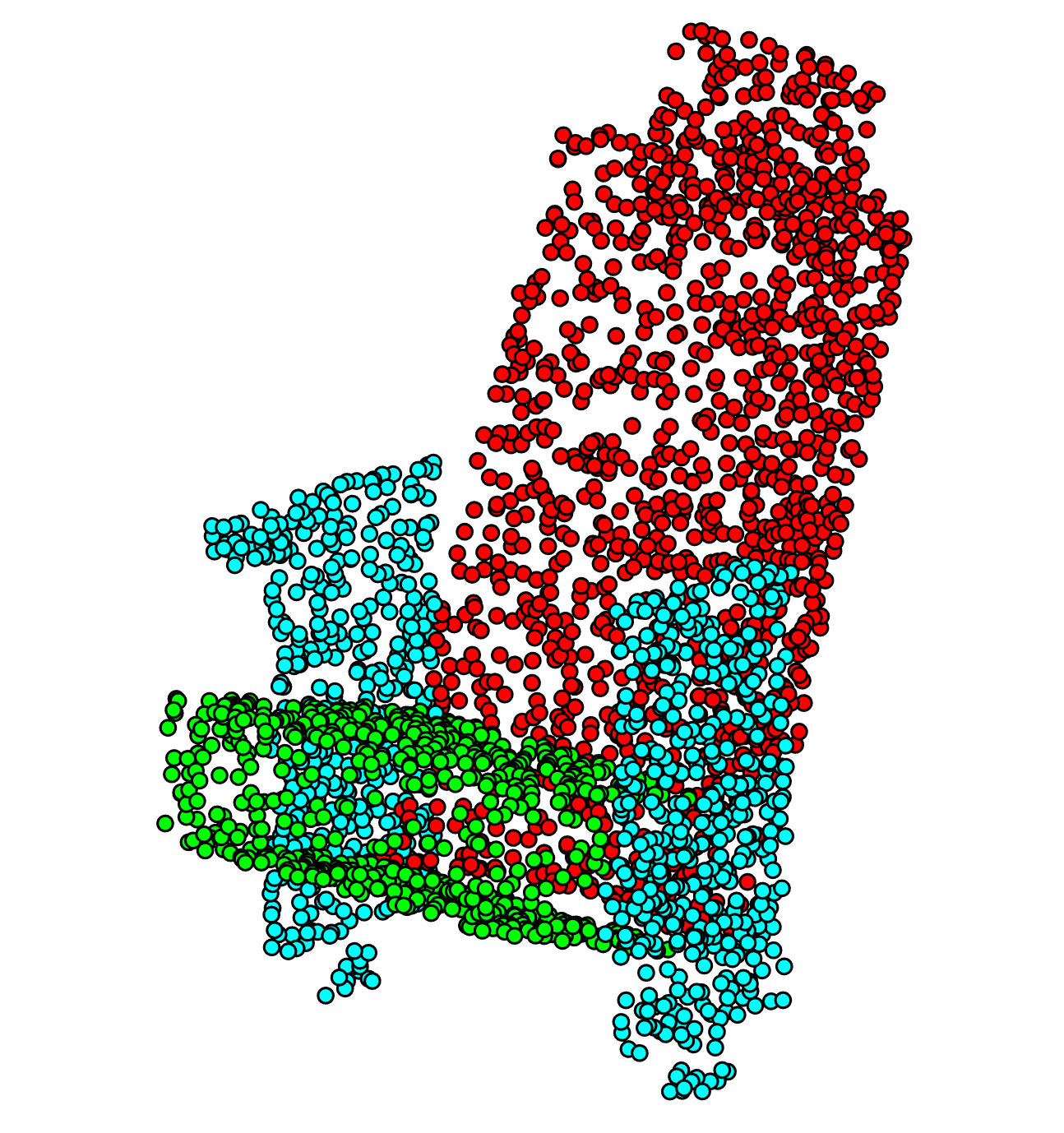}   &
    \includegraphics[width=16mm,height=17mm]{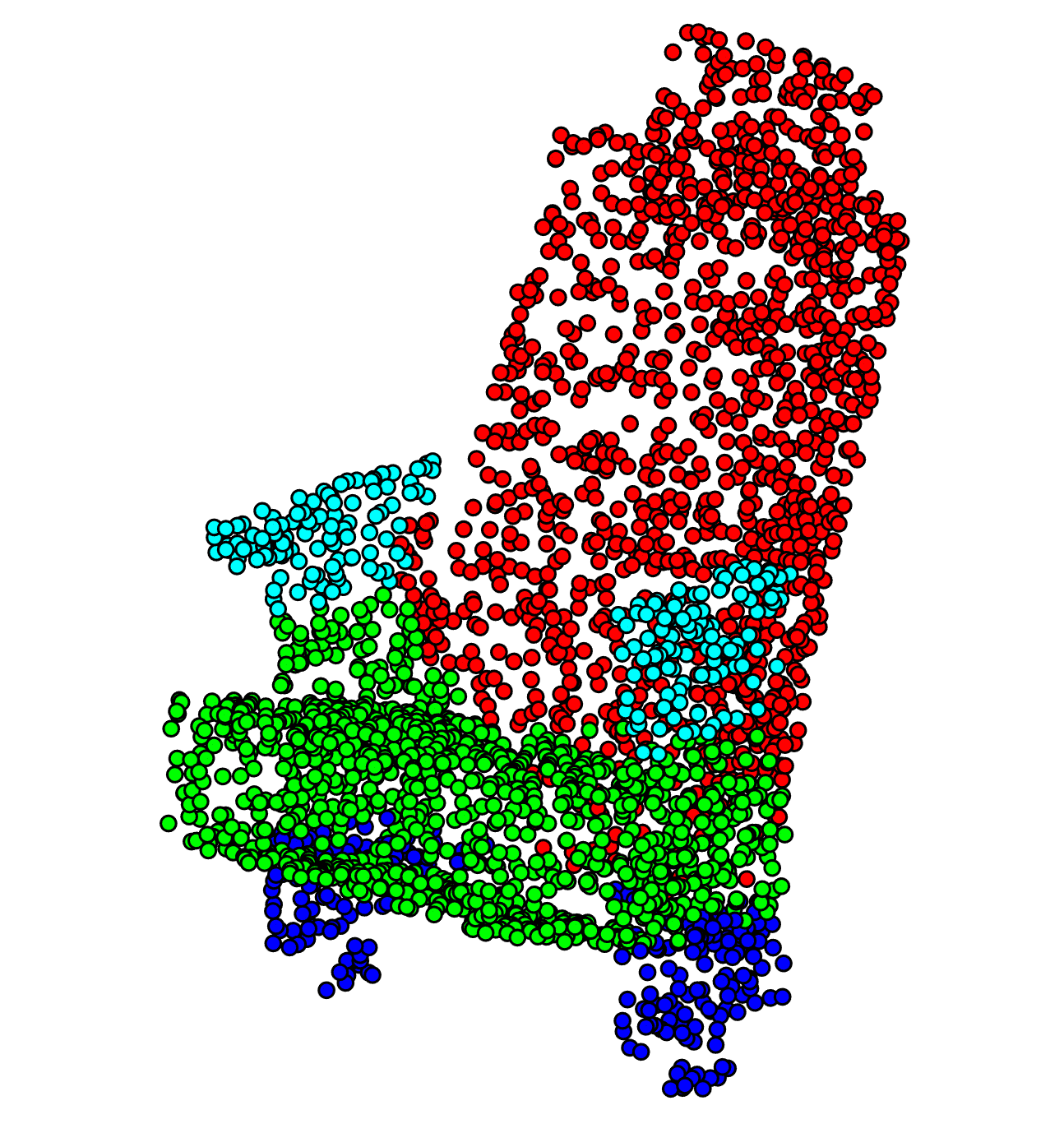}  \\
   Skateboard & 92.2\% & Chair &96.0\% & Skateboard&55.8\% & Chair &41.6\% \\
   \hline
   \end{tabular}
  \caption{Representative examples of high- and low-quality segmentation results of $\Psi$-CNN. Computed mIoU is also given in each case. Low-quality segemetation generally result from: (1) confusing ground truth labeling, e.g.~axles of \textit{skateboards} are considered separate segments in most of the ground-truth labels, (2)  small object parts with no clear boundaries, e.g.~handles of \textit{bags}. Color coding is within category (best viewed on screen).
%  It can be noticed that among the low-quality segmentations,
%some poor predictions are caused because the parts are too delicate to be learned correctly, like the small handles of the \emph{bag} and the tiny lights of the \emph{lamp}. Besides, other poor predictions are mostly caused by the confusing labelling cast by the ground truth. For example, in the ground
%truth of \emph{skateboard}, the axles are sometimes labelled separately from the wheels and sometimes not. Similarly, in the \emph{Chair}
%category, the back and legs are sometimes labelled as separate parts and sometimes not. For networks which make use of $xyz$ coordinates information only, such ambiguous ground truth labelling confuses the network, and it tends to make predictions based on the dominant cases in each category only.
  }\label{partseg_examples}
\end{figure*}

\begin{table*}
\caption{Results on ShapeNet part segmentation dataset}\label{table:partSeg}
\begin{adjustbox}{width=1\textwidth}
{\LARGE\begin{tabular}{l|c|c|c|c|c|c|c|c|c|c|c|c|c|c|c|c|c|c}
  \hline
 Method & mIoU & NO. & Airplane & Bag & Cap & Car & Chair & Earphone & Guitar & Knife & Lamp & Laptop & Motorbike & Mug & Pistol & Rocket & Skateboard & Table \\
  \hline
%  \# shapes & & &2690& 76& 55& 898& 3758 &69 &787 &392 &1547 &451 &202 &184 &283 &66 &152 &5271\\
%   \hline
 % \hline
  3D-CNN~\cite{qi2017pointnet}& 79.4&0 & 75.1 &72.8 &73.3 &70.0 &87.2 &63.5 &88.4 &79.6 &74.4 &93.9 &58.7 &91.8 &76.4 &51.2 &65.3 &77.1\\
%   O-CNN+CRF\cite{wang2017cnn} & 85.9 &85.5 &87.1 &84.7 &77.0 &91.1 &85.1 &91.9 &87.4 &83.3 &95.4 &56.9 &96.2 &81.6 &53.5 &74.1 &84.4 \\
  \hline
  Kd-net~\cite{klokov2017escape}
& 82.3 &0&80.1 &74.6 &74.3 &70.3 &88.6 &73.5 &90.2 &87.2 &81.0 &94.9 &57.4 &86.7 &78.1 &51.8 &69.9 &80.3 \\
  PointNet~\cite{qi2017pointnet}&  83.7&0  &83.4 &78.7 &82.5 &74.9 &89.6 &73.0 &91.5 &85.9 &80.8 &95.3 &65.2 &93.0 &81.2 &57.9 &72.8 &80.6 \\
  SyncSpecCNN~\cite{yi2017syncspeccnn}& 84.7 &2 &81.6 &81.7 &81.9 &75.2 &90.2 &74.9 &\textbf{93.0} &86.1 &84.7 &95.6 &66.7 &92.7 &81.6 &60.6 &\textbf{82.9} &82.1\\
  KCNet~\cite{shen2018mining}& 84.7&1 &82.8 &81.5 &86.4 &77.6& 90.3 &76.8 &91.0 &\textbf{87.2} &84.5 &95.5 &69.2 &94.4 &81.6 &60.1 &75.2 &81.3\\
 SO-Net~\cite{li2018so}& 84.9 &1 &82.8 &77.8 &\textbf{88.0} &77.3 &90.6 &73.5 &90.7 &83.9 &82.8 &94.8 &69.1 &94.2 &80.9 &53.1 &72.9 &83.0 \\  PointNet++~\cite{qi2017pointnetplusplus} & 85.1 &0&82.4 &79.0 &87.7 &77.3 &90.8 &71.8 &91.0 &85.9 &83.7 &95.3 &71.6 &94.1 &81.3 &58.7 &76.4 &82.6 \\
%   SpiderCNN &85.3 &83.5 &81.0 &87.2 &77.5 &90.7& 76.8& 91.1 &87.3 &83.3 &95.8 &70.2 &93.5 &82.7 &59.7 &75.8 &82.8\\
%   $\Psi$-CNN (proposed) & \textbf{87.5}(80.31)
%   & \textbf{86.8}(76/77) &\textbf{88.8}(64/73) &\textbf{90.2}(61) &\textbf{80.3}(69) &\textbf{92.6}(87) &\textbf{86.4}(59) &\textbf{93.7}(87) &\textbf{87.9}(83) &\textbf{86.8}(78) &\textbf{98.8}(94) &\textbf{73.5}(50) &\textbf{96.5}(90) &\textbf{86.2}(74) &\textbf{68.5}(45) &\textbf{80.2}(65) &\textbf{86.5}(79) \\
\hline
% $\Psi$-CNN &\textbf{83.0}& \textbf{85.5}
% &\textbf{84.0} &\textbf{82.1} &83.8 &\textbf{80.0} &\textbf{90.9} &76.2 &91.6 &86.7 &84.3 &\textbf{95.6} & \textbf{74.8} &\textbf{94.5} & \textbf{83.4} & \textbf{61.3} &75.9 &82.2\\
$\Psi$-CNN &%\textbf{83.4}&
\textbf{86.8}&\textbf{11}
&\textbf{84.2} &\textbf{82.1} &83.8 &\textbf{80.5} &\textbf{91.0} &\textbf{78.3} &91.6 &86.7 &\textbf{84.7} &\textbf{95.6} & \textbf{74.8} &\textbf{94.5} & \textbf{83.4} & \textbf{61.3} &75.9 &\textbf{85.9}\\
  \hline
\end{tabular}}
\end{adjustbox}
\end{table*}
\subsection{Semantic Segmentation}
\vspace{-2mm}
We also test our model for Semantic Segmentation of real world data with RueMonge2014 dataset~\cite{riemenschneider2014learning}.
This dataset contains 700 meter facades along a street annotated with point-wise labelling. The classes includes \emph{window, wall, balcony, door, roof, sky} and \emph{shop}. The point clouds are provided with color features. To train our network, we split both the training and testing data into $1m^3$ blocks. We align the facade plane of all the blocks into the same plane, and adjust the gravitational axis to be upright. We only force the $x$ and $y$ dimensions to have zero-means, but not the $z$ axis. This processing strategy is adopted to avoid loosing the height information. 
We use $xyz$+$rgb$ as input raw features to train our network. The used network configuration is MLP(64)-Octree(64-64-128-128-256-256). Table \ref{ruemonge2014_seg} compares the results of our approach with 
the current state-of-the-art on this dataset, 
under the evaluation protocol of \cite{gadde2018efficient}. With 7 parametric layers, we achieve better performance than OctNet,
which uses 20 parametric layers to learn the final representation of each point. These results demonstrate the promises of $\Psi$-CNN in practical applications. %Visualizations for the segmentation results are provided in the supplementary material.

\begin{table}
\caption{Semantic Segmentation on RueMonge2014 dataset}\label{ruemonge2014_seg}
\begin{tabular}{l|c|c|c}
  \hline
 Method& Average & Overall & IoU \\
  \hline
Riemenschneider et al. \cite{riemenschneider2014learning} & -- & -- & 42.3 \\
Martinovic et al. \cite{martinovic20153d} & -- & -- & 52.2 \\
Gadde et al. \cite{gadde2018efficient} & 68.5 &78.6 &54.4 \\
%OctNet $64^3$ \cite{riegler2016octnet} & 60.0 & 73.6 & 45.6\\
%OctNet $128^3$ \cite{riegler2016octnet} & 65.3 & 76.1 & 50.4 \\
OctNet $256^3$ \cite{riegler2017octnet} & 73.6 & 81.5 & 59.2 \\
\hline
$\Psi$-CNN & \textbf{74.7} & \textbf{83.5} & \textbf{63.6} \\
  \hline
\end{tabular}\
\end{table}

\subsection{Discussion}
%\begin{figure}[!t]
%\includegraphics[width=25mm]{octree-knn-1.png}
%\hspace{1mm}
%\includegraphics[width=25mm]{octree-range-1.png}
%\hspace{1mm}
%\includegraphics[width=25mm]{octree-kdtree-1.png}
%\caption{Comparison of octree structuring with K-NN, range search and Kd-tree for neighborhood computation. %}\label{time-analysis}
%\label{fig:graphs}
%\end{figure}
\begin{figure}[h!]
\includegraphics[width=\columnwidth]{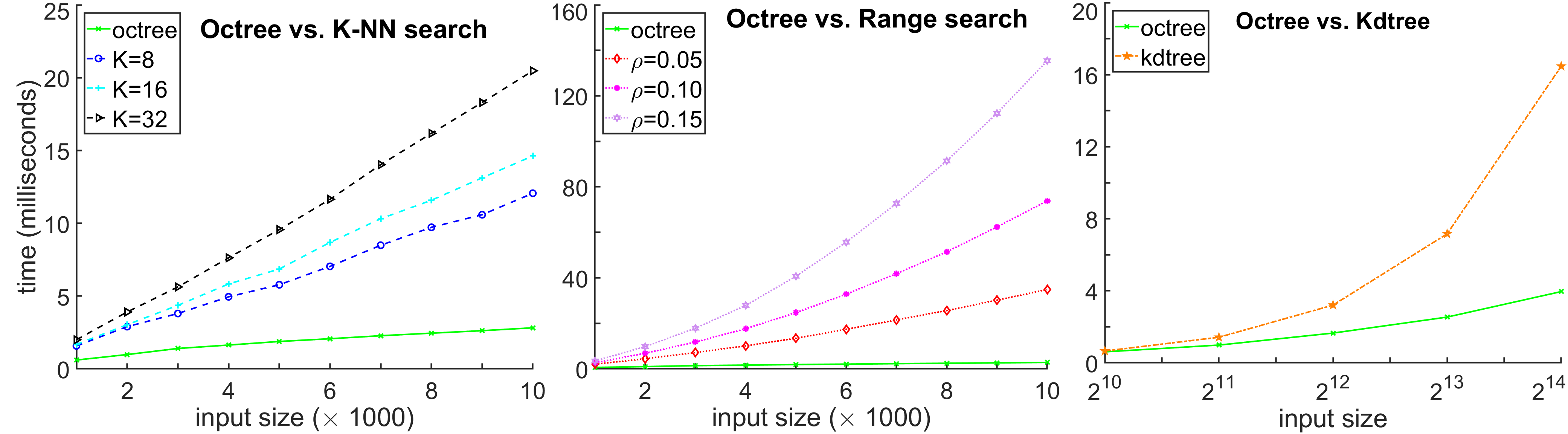}
\caption{Comparison of octree structuring with K-NN, range search and Kd-tree for neighborhood computation. }\label{time-analysis}
\label{fig:graphs}
\vspace{-4mm}
\end{figure}
For geometrically meaningful convolutions, knowledge of local neighborhood of points is imperative. A related approach, ECC \cite{simonovsky2017dynamic} exploits range search for this purpose. Another obvious choice is K-NN clustering. However, with tree structures, e.g.~octree; the point neighborhood information is already readily available that adds to computational efficiency of $\Psi$-CNN. In Fig.~\ref{fig:graphs}, we report the timings of computing neighborhoods under different choices, and compare them to octree construction. As can be seen, for larger number of input points, octree structuring is more efficient as compared to K-NN and range searching. Moreover, its efficiency is also better than Kd-tree for large input sizes because the binary split in Kd-tree forces it to be much deeper than octree.

Running our classification network on 1K randomly selected samples from ModelNets, we compute the test time of our network for point clouds of sizes 10K, and report timings in Table \ref{tab:test_time}. 
The test time for a sample consists of time required to construct the octree and performing the forward pass. We also show the time of normal computation in the table \textit{for reference}.  Our approach does not compute normals to achieve the results reported in the previous section. To put these timings into perspective,  PointNet++ \cite{qi2017pointnetplusplus} requires roughly 115ms for a forward pass of input with 1024 points on the same machine. In Fig.~\ref{fig:octree_coarsen}, we also show a representative example of point cloud coarsening by our method under octree structuring. Our network gradually sparsifies the point cloud by applying spherical convolutional kernel at each level. %{\color{red}Visualizations of the learned spherical kernels are provided in the supplementary material.}
\begin{table}
\centering
\begin{adjustbox}{width=0.47\textwidth}
{\LARGE\begin{tabular}{c|c|c|c||c}
  \hline  
 Input size & Octree construction & Forward pass&~~~~~Total~~~~~& \cellcolor{red!15} Normal computation \\
  \hline
  \hline
  10K & 3.5 & 30.6 & 34.1& \cellcolor{red!7}27.4\\
  \hline
\end{tabular}}
\end{adjustbox}
\caption{Per-sample test time (ms) for 10K input. The computing time for normals is included for reference only - indicated by red.}
\label{tab:test_time}
\vspace{-2mm}
\end{table}
\begin{figure}
    \centering
    \begin{tabular}{ccccc}
         \includegraphics[width=13mm]{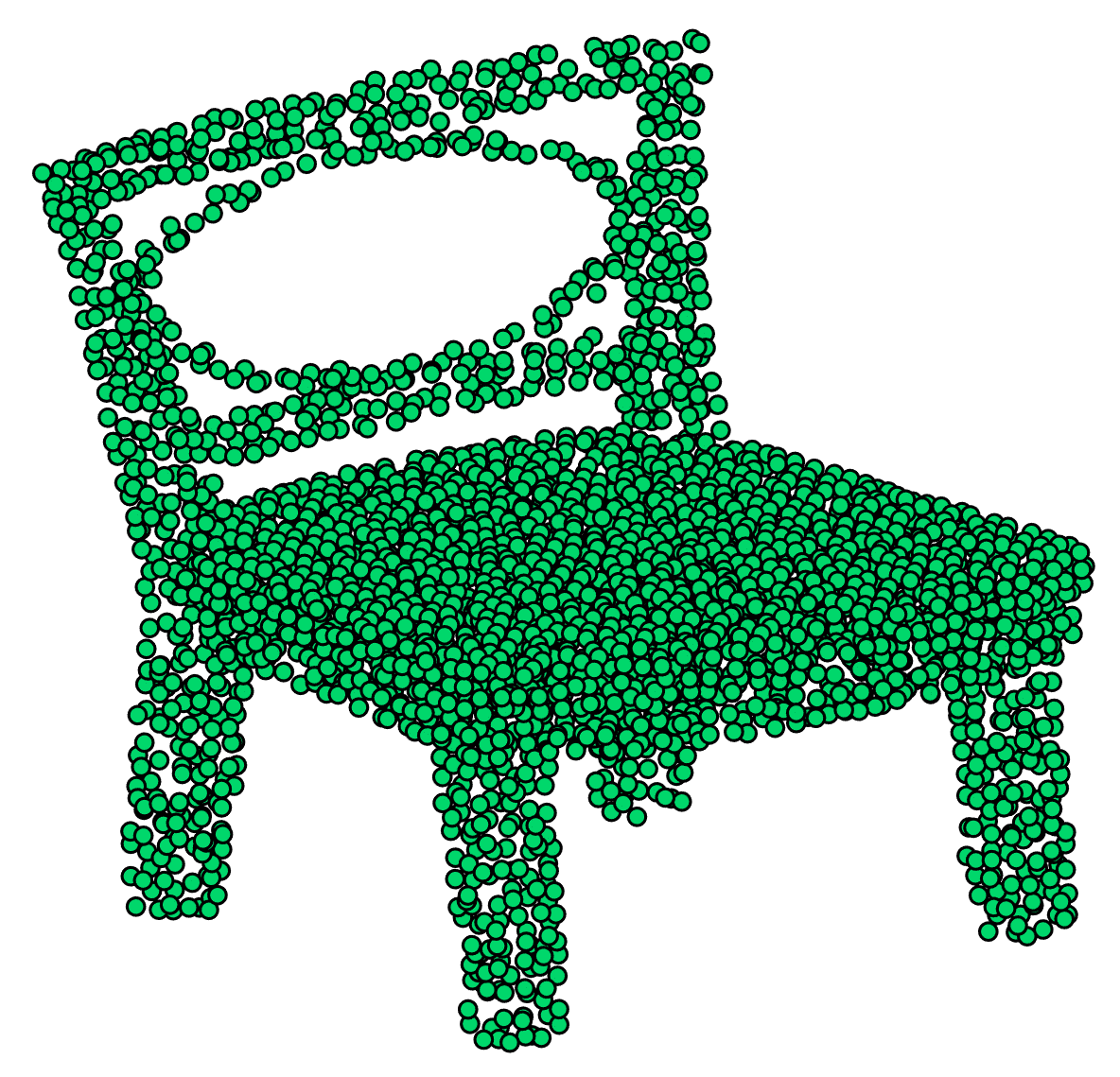}&  
         \includegraphics[width=13mm]{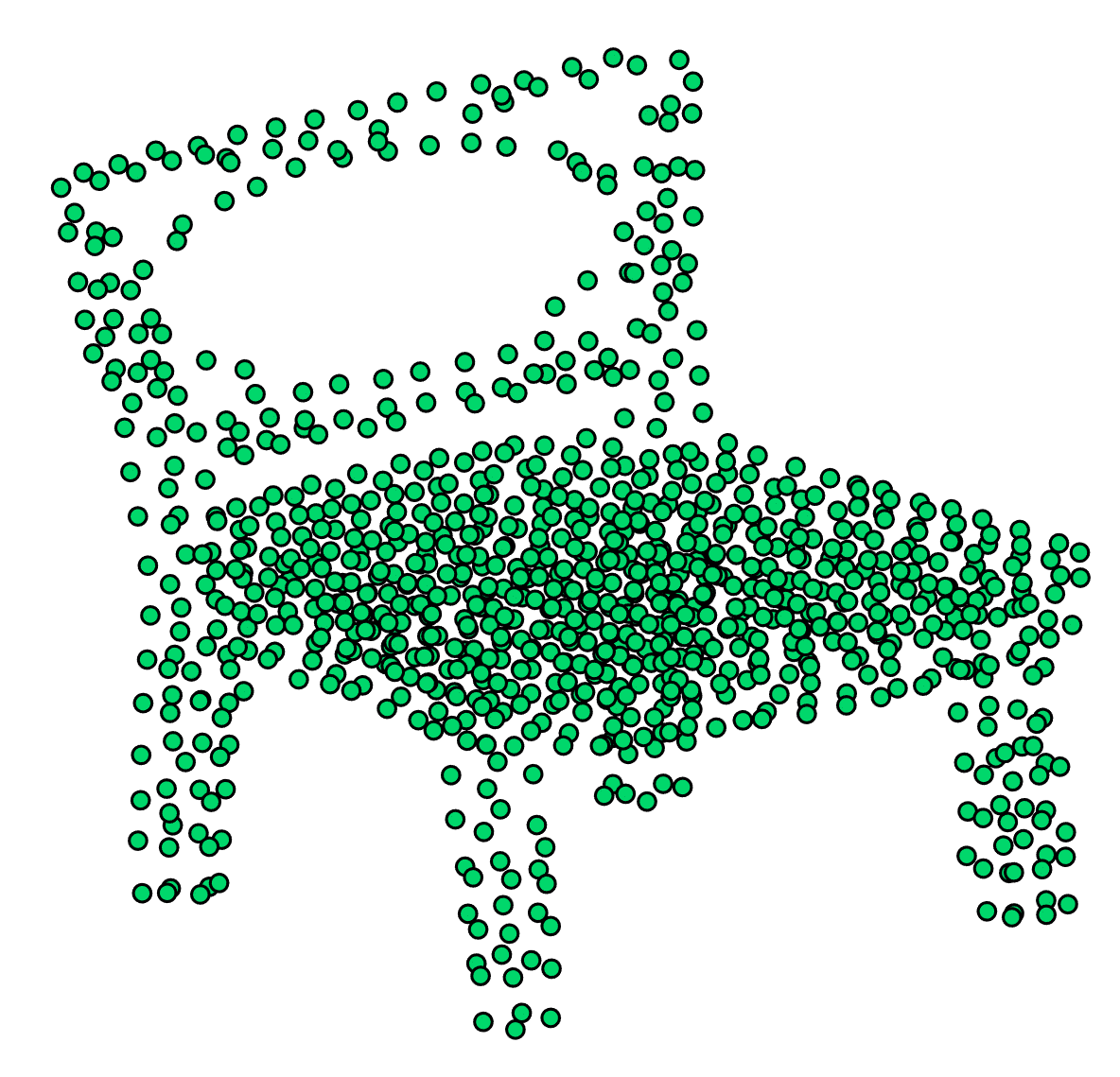}& 
         \includegraphics[width=13mm]{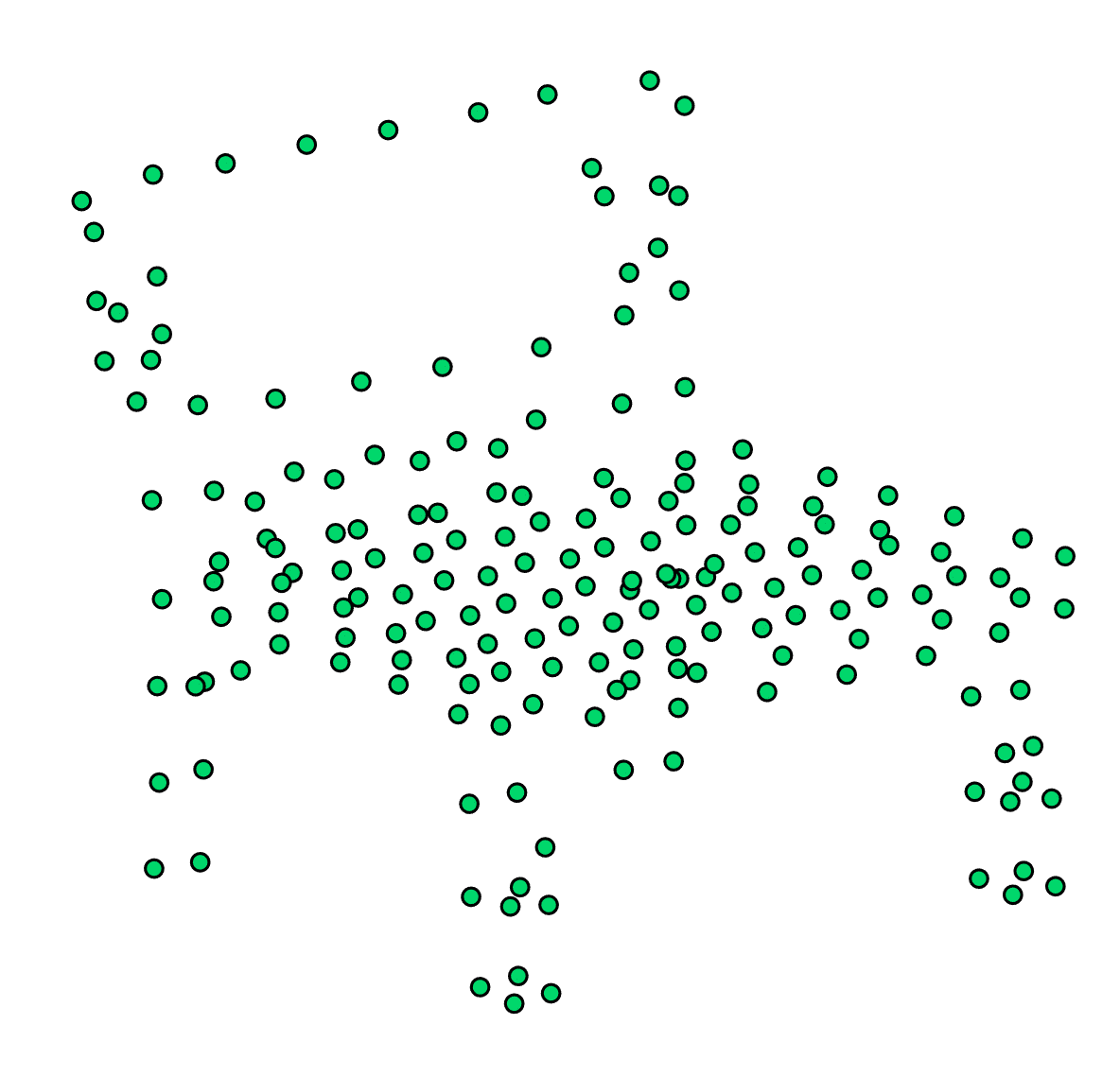}& 
         \includegraphics[width=13mm]{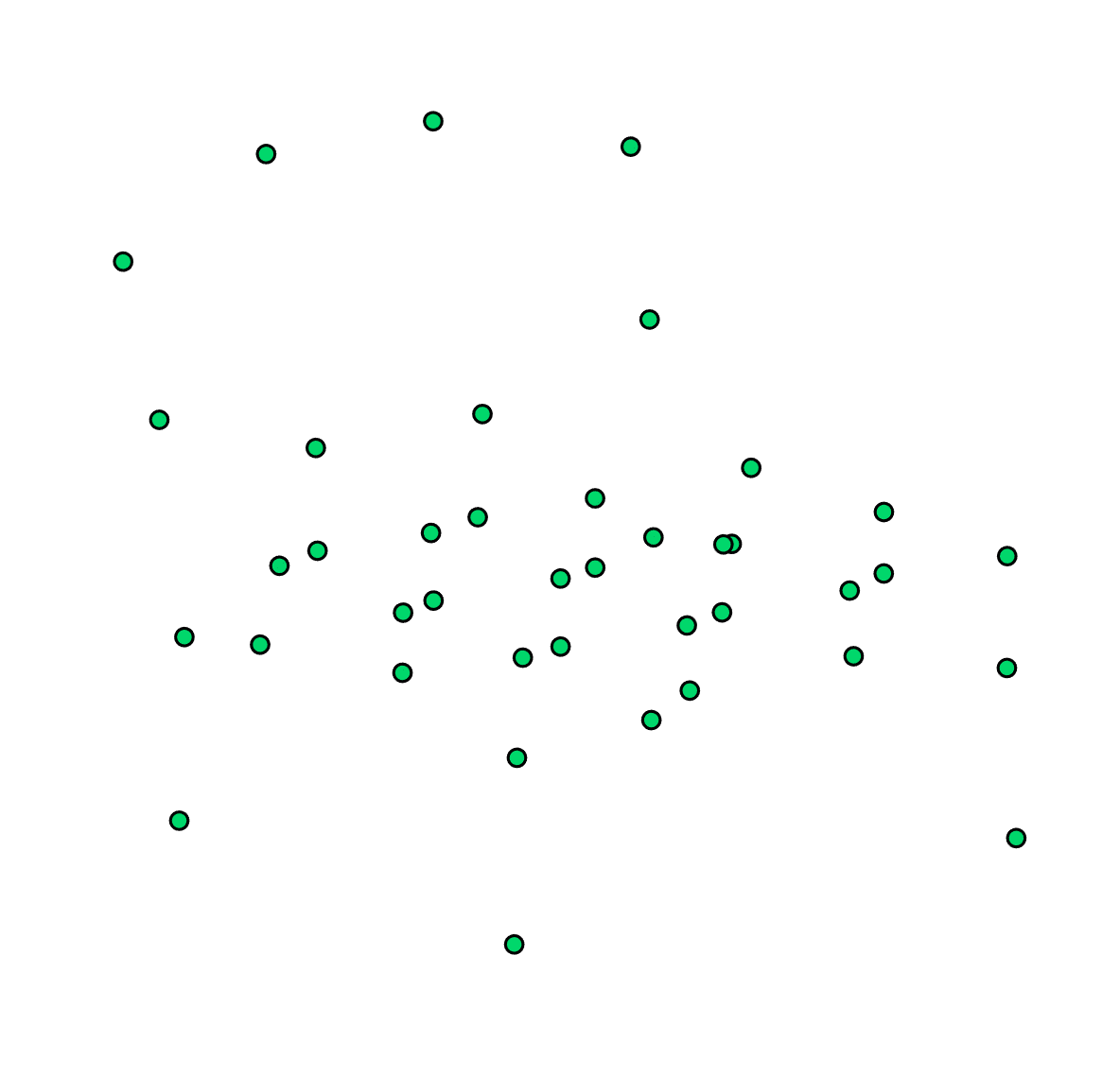}& 
         \includegraphics[width=13mm]{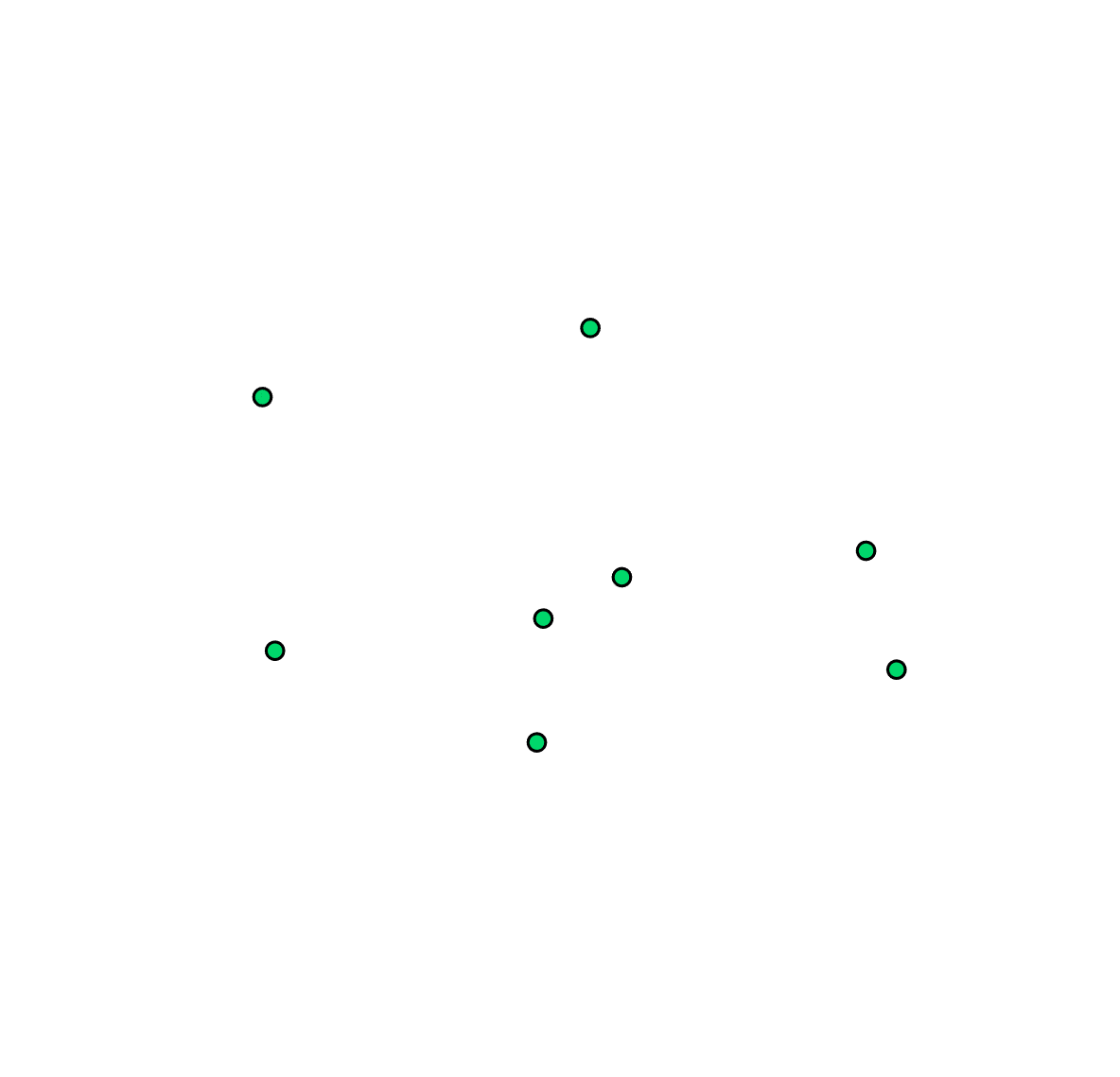}\\
         $l=1$ & $l=2$ & $l=3$ & $l=4$& $l=5$ \\
    \end{tabular}
    \caption{Point cloud coarsening example under octree structuring by our technique. `$l$' is the octree level.}
    \label{fig:octree_coarsen}
    \vspace{-5mm}
\end{figure}
\section{Conclusion}
We introduced the notion of spherical convolutional kernels for point cloud processing and demonstrated its utility with a neural network guided by octree structure. The network successively performs convolutions in the neighborhood of its neurons, the locations of which are governed by the nodes of the underlying octree. To perform the convolutions, our  spherical kernel divides its occupied space into multiple bins and associates a weight matrix to each bin. These matrices are learned with network training. We have shown that the resulting network can efficiently process large 3D point clouds in effectively achieving excellent performance on the tasks of 3D classification and segmentation on synthetic and real data. %Other hierarchical structures that facilitate the spherical kernel will be explored in the future.

{\noindent \textbf{Acknowledgments}} This research was supported by
ARC Discovery Grant DP160101458 and partially by
DP190102443. We also thank NVIDIA corporation for donating
the Titan XP GPU used in our experiments.

\end{document}